\documentclass[12pt]{article}

\usepackage[a4paper,margin=2.5cm]{geometry}
\usepackage{setspace}
\usepackage{graphicx}
\usepackage{booktabs}
\usepackage{tabularx}
\usepackage{array}
\usepackage{longtable}
\usepackage{ragged2e}
\usepackage{amsmath}

\usepackage[a4paper,margin=2.5cm]{geometry}
\usepackage{enumitem}
\usepackage[T1]{fontenc}
\usepackage{lmodern}

\newcolumntype{L}[1]{>{\RaggedRight\arraybackslash}p{#1}}
\newcolumntype{Y}{>{\RaggedRight\arraybackslash}X}

\usepackage[round,authoryear]{natbib}
\usepackage{xurl}
\usepackage[hidelinks]{hyperref}

\usepackage[T1]{fontenc}
\usepackage[utf8]{inputenc}

\begin{document}


\title{
Evaluating human and LLM screening workflows in a conceptually complex scoping review: Recall--workload trade-offs and run-to-run consistency
}

\author{
Nikol Figalová\textsuperscript{1},
Lynn Huestegge\textsuperscript{1,*},
and Anne Böckler-Raettig\textsuperscript{1,*}
\\[1em]
\textsuperscript{1}Institute of Psychology,
Julius-Maximilians-Universität Würzburg,\\
Würzburg, Germany
\\[0.75em]
\textsuperscript{*}These authors contributed equally
\\[1em]
\textbf{Corresponding author:}\\
Nikol Figalová\\
Institute of Psychology,
Julius-Maximilians-Universität Würzburg\\
Würzburg, Germany\\
\href{mailto:nikol.figalova@uni-wuerzburg.de}
{nikol.figalova@uni-wuerzburg.de}
}

\date{}

\maketitle

\clearpage


\begin{abstract}

\noindent\textbf{Background.}
Large language models (LLMs) are increasingly used for document-classification tasks in evidence synthesis, where false-negative decisions can remove relevant studies before full-text assessment. We evaluated human and LLM-based title-and-abstract screening workflows in a preregistered methodological study embedded in a conceptually complex scoping review, treating the implemented workflow rather than the model alone as the primary unit of comparison.

\medskip
\noindent\textbf{Methods.}
After a conservative preliminary title-only screen, 1,131 records were screened by one review lead, a distributed team of four trained assistants screening non-overlapping subsets, and seven complete LLM runs spanning different models and processing configurations, including a nominally identical repeat run. We compared retained workload, operational recall against 316 verified eligible records, conditional classification performance among 859 full-text-assessed records, agreement, combined recovery, run-to-run consistency, and procedural burden. Because eligibility was verified only for records advanced and assessed in the parent review, recall estimates were operational rather than complete-benchmark estimates.

\medskip
\noindent\textbf{Results.}
No individual workflow recovered all verified eligible records. The two human workflows and two GPT-5.4 file-batch runs showed similar workload--recall profiles, retaining 42.2--45.0\% of records while achieving operational recall of 82.3--82.9\%. Gemini 3.1 file batches achieved the highest recall (83.9\%) but retained 56.7\% of records. Both all-at-once configurations recovered fewer eligible records than their corresponding file-batch configurations. Two nominally identical GPT-5.4 file-batch runs agreed on 91.7\% of records, yet differed on 94 records, including 29 verified eligible records retained by only one run.

\medskip
\noindent\textbf{Discussion.}
LLM screening performance depended on the complete implemented workflow and could not be characterised by model identity or aggregate performance alone. Processing configuration, downstream workload, record-level variation, and the way model outputs are combined with human decisions are substantive properties of the deployed system. For high-recall screening tasks, LLMs are better suited to validated, auditable, human-supervised workflows than to autonomous exclusion.

\end{abstract}


\medskip


\medskip

\clearpage



\section{Introduction}

Large language models (LLMs) are increasingly being used as natural-language classifiers within decision-support workflows, including tasks in which errors have asymmetric consequences. Title-and-abstract screening in evidence synthesis provides a demanding real-world example: large numbers of short documents must be classified against natural-language eligibility criteria, while false-negative decisions can permanently remove relevant evidence from subsequent assessment. At the same time, overly conservative classification increases the number of records requiring full-text retrieval and review. Screening therefore involves a practical trade-off between recovering relevant evidence and limiting downstream workload \citep{homiar2025development,landschaft2024implementation,shailendra2025lprisma}.

This trade-off is particularly consequential in scoping reviews. Such reviews often address broad or interdisciplinary questions, include heterogeneous study designs, and require judgements about conceptual relevance rather than application of narrowly specified eligibility criteria \citep{peters2020updated,tricco2018prisma}. Conceptual and terminological heterogeneity may also create jingle and jangle problems, whereby distinct phenomena share a label or closely related phenomena are described using different terminology \citep{hanfstingl2024detecting}. Reviewers must consequently distinguish clearly relevant and irrelevant records from those for which the title and abstract provide insufficient information. Screening is commonly conservative in such cases, with ambiguous records retained for full-text assessment.

LLMs offer a potentially useful means of supporting this process because eligibility criteria can be expressed directly in natural language and applied to titles and abstracts without review-specific model training. However, the relevant unit of evaluation is not necessarily the underlying model alone. An implemented LLM screening system also includes the prompt, uncertainty rule, record preparation, processing configuration, interaction procedure, output format, quality-control steps, and the rule by which model outputs are translated into downstream decisions. We therefore treat the \emph{screening workflow}, rather than model identity in isolation, as the primary unit of comparison.

\subsection{LLM-based screening as a workflow-level problem}

Earlier approaches to automating evidence synthesis predominantly relied on conventional machine-learning and active-learning methods. Tools such as ASReview use reviewer-labelled examples to prioritise records for screening, reducing the number of irrelevant records humans must inspect while retaining human oversight \citep{vandeschoot2021asreview}. These approaches generally depend on reviewer-generated labels, iterative learning, or corpus-specific calibration.

LLMs enable a different form of automation. Through natural-language prompting, they can interpret eligibility criteria and return screening decisions without review-specific model training. LLMs are now being investigated across evidence-synthesis tasks including title-and-abstract screening, full-text assessment, data extraction, and synthesis \citep{harasgama2026artificial,lieberum2025large,njei2026artificial,cao2026automation,laignelot2026large}. Recent reviews nevertheless recommend restricting their use to clearly defined tasks, transparently documenting the implemented procedure, validating performance in the intended context, and maintaining appropriate human supervision \citep{lieberum2025large,galli2025large,laignelot2026large}.

Empirical evaluations show that LLM screening performance varies across models, prompts, datasets, review topics, decision thresholds, and reference standards. GPTscreenR, for example, used GPT-4 for title-and-abstract screening in scoping reviews and demonstrated useful but imperfect performance against consensus human decisions, with human--GPT agreement below inter-human agreement \citep{wilkins2023automated}. Other comparative studies have similarly reported substantial variation across model families, prompting strategies, and review datasets \citep{li2024evaluating,syriani2024screening,huotala2024promise,krag2024large,dennstadt2024title}. Even relatively small changes to instructions, response scales, or classification thresholds can alter the operating characteristics of the resulting screening procedure \citep{dennstadt2024title}.

Processing configuration may also be consequential. The number of records submitted within a prompt can affect both processing feasibility and classification performance, indicating that batch size may be a substantive workflow parameter rather than a neutral implementation choice \citep{fagerberg2026batch}. More elaborate procedures do not necessarily improve performance. Akinseloyin et al. \citep{akinseloyin2026multiagent}, for example, found that averaging independently generated relevance scores from several lower-cost LLMs provided robust and cost-effective prioritisation, whereas more complex debate and LLM-based adjudication procedures did not consistently add benefit.

Repeated execution introduces a further source of variation. LLM outputs may differ even when the model, records, and instructions appear unchanged \citep{syriani2024screening}. Similar aggregate recall or workload across repeated runs therefore does not imply that the same individual records were classified identically. From a workflow-evaluation perspective, reproducibility must consequently be assessed not only through aggregate performance but also through record-level agreement and the identities of discordantly classified records.

These considerations make attribution of performance to a model name alone problematic. Reporting frameworks for LLM-assisted evidence synthesis accordingly emphasise documentation of the model and version, prompts, input preparation, processing configuration, decision rules, human involvement, validation procedures, and reference standards \citep{susnjak2023prisma,shailendra2025lprisma,holst2025prismatraice,gallifant2025tripodllm,luo2025gamer}. Such information is necessary to characterise the implemented system and to understand which procedural choices may have contributed to observed performance.

\subsection{Evaluating screening performance}

No single metric fully characterises a screening workflow. Agreement with human title-and-abstract decisions, recovery of records ultimately judged eligible after full-text assessment, and reduction of downstream full-text workload capture related but distinct properties. An LLM may agree closely with human screeners while nevertheless missing records later found to be eligible. Conversely, high recovery may be achieved simply by retaining a large proportion of the screened dataset. Evaluations should therefore report recovered and missed evidence alongside the number of records retained for subsequent assessment rather than relying on a single summary measure \citep{homiar2025development,landschaft2024implementation,sciurti2026compact,madeyski2025llm4screenlit}.

Agreement provides complementary information but is not equivalent to screening performance. Because most records retrieved through review searches are typically ineligible, high overall agreement can be driven by concordant exclusions even when workflows differ on a smaller but consequential group of potentially eligible records. Conversely, workflows may use categories such as \emph{Include} and \emph{Unclear} differently while retaining largely overlapping sets of records. Agreement can therefore be examined both at the level of the original decision categories and at the operational retained/not-retained level, depending on the evaluation objective.

Reference-based classification measures introduce an additional methodological constraint \citep{sokolova2009systematic}. Measures such as recall, precision, specificity, and the F$_1$ score require verified reference outcomes for the records included in their calculation. In evidence synthesis, however, final eligibility is usually established only for records that pass title-and-abstract screening and proceed to full-text assessment. The reference set is therefore partly determined by the screening workflows themselves rather than by independent verification of the complete dataset.

This creates a form of partial verification bias \citep{whiting2011quadas}. A record can receive a verified final eligibility outcome only if it is retained by at least one advancement workflow, its full text is successfully retrieved, and it undergoes full-text assessment. Records excluded by all advancement workflows, or those for which full texts cannot be retrieved, have unknown final eligibility and cannot automatically be treated as true negatives. Consequently, recall estimates based on subsequently verified eligible records are operational estimates rather than complete-benchmark measures, while other reference-based metrics apply only to the subset with verified full-text outcomes.

The outputs of different workflows may also be complementary. Under a liberal combination rule, retaining a record when either member of a pair recommends advancement can reduce jointly missed eligible records, although generally at the cost of greater downstream workload. Large-scale evaluations have shown that human--LLM and LLM--LLM combinations can improve sensitivity, with the benefit depending on the review and the chosen combination rule \citep{sanghera2025high}. More conservative approaches have instead accepted automated decisions only when multiple LLMs agree and referred discordant cases for human assessment \citep{hilkenmeier2026full}. Evaluation of an AI-assisted screening system must therefore consider not only individual classifiers but also how their outputs are integrated into the final decision process.

Human decisions themselves are not an error-free reference standard \citep{wang2020error}. The number and organisation of reviewers can affect both reliability and resource requirements, and single-reviewer screening cannot be assumed to recover the same records as screening involving additional reviewers \citep{waffenschmidt2019single}. Human reviewers may differ in expertise, interpretation of eligibility criteria, calibration, and tolerance for uncertainty, while performance can also be influenced by task framing and fatigue \citep{belur2021interrater,wang2020error}. Human judgement under uncertainty is more generally susceptible to systematic heuristics and biases \citep{tversky1974judgment}.

LLM errors arise through different mechanisms. A model may misunderstand an abstract, apply a criterion too strictly or too leniently, over-rely on salient topical terms, or fail to follow instructions for handling ambiguity. Such errors are consistent with broader evidence that LLM failures may reflect limitations in knowledge, reasoning, and instruction following \citep{wu2026prism}. Comparing realistic human and LLM workflows is therefore more informative than treating either as an idealised or error-free decision maker.

Despite the rapidly expanding literature on LLM-assisted evidence synthesis, several gaps remain. Existing evaluations are concentrated in biomedical or intervention-focused systematic reviews with comparatively structured eligibility criteria, leaving less evidence from conceptually complex and interdisciplinary screening tasks. Relatively few studies compare LLM procedures with both a solo review lead and a distributed trained human team using the same records, criteria, decision categories, and advancement rule. Processing configuration and repeated execution also remain insufficiently characterised under realistic web-interface conditions. More broadly, evaluations frequently report model-level performance without fully separating recovery, downstream workload, agreement, conditional classification performance, record-level reproducibility, and complementarity among different decision workflows.

\subsection{The present study}

The present study was a preregistered comparative methodological evaluation embedded within a PRISMA-ScR-aligned scoping review of gaze semantics in human social interaction \citep{figalova_bockler_huestegge_2026}. The parent review examined research in which gaze behaviours were treated as meaningful, interpretable, communicative, or socially consequential in human social contexts.

This application provides a challenging natural-language classification task because eligibility depends on conceptual interpretation rather than simple topical matching. A study may mention gaze or eye tracking without examining the social meaning of gaze, whereas another may investigate social inference or communication without describing its contribution explicitly as \textit{gaze semantics}. Screening therefore requires interpretation of how the phenomenon is conceptualised and whether the information provided in the title and abstract is sufficient for a confident decision.

We compared screening by a solo review lead, a distributed team of four trained psychology student assistants, and seven complete LLM runs spanning multiple models and processing configurations. All workflows were evaluated on the same benchmark dataset and under a common operational retention rule. Rather than treating model identity as the sole unit of comparison, we evaluated the complete implemented workflows, including differences in how records were submitted to the LLM and whether nominally identical procedures produced stable record-level decisions across repeated execution.

We use \emph{workflow} to denote a complete screening procedure defined by the screening entity, model where applicable, input format, prompting strategy, and output-handling process. A \emph{run} is one complete execution of a workflow on the benchmark records. \emph{Processing configuration} describes how records were submitted to the LLM, for example as one complete file, separate file batches, or smaller sequential batches. We further distinguish \emph{advancement workflows}, whose decisions contributed to the records retrieved for full-text assessment in the parent review, from \emph{comparative runs}, which were evaluated retrospectively and did not influence retrieval.

The evaluation deliberately separated several dimensions of performance: recovery of verified eligible records, retained full-text workload, agreement among workflows, conditional reference-based classification performance among records with verified full-text outcomes, and run-to-run consistency. Record-level analyses additionally examined overlap in missed eligible records, unique recovery, and the consequences of combining pairs of screening outputs under a liberal advancement rule. This design allowed us to examine not only whether LLM-based screening could approximate human performance, but also whether performance depended on processing configuration, whether repeated executions produced the same decisions, and whether human and LLM outputs provided complementary information.

The study addressed the following research questions:

\begin{enumerate}
\item How did the human and LLM-based screening workflows differ in their recovery of verified eligible records and in the number of records retained for full-text assessment?
\item To what extent did the screening decisions produced by the different workflows agree?
\item How did the workflows perform on conditional reference-based classification measures among records assessed at full text?
\item How consistent were two LLM screening runs conducted under nominally identical conditions?
\item How did the workflows differ in the verified eligible records they failed to retain, and what did retrospective checks reveal about records retained only by a later comparative LLM run?
\end{enumerate}

Screening time, interaction burden, and other procedural characteristics were additionally summarised where corresponding data were available. Pairwise combined recovery was examined as an exploratory analysis.

\section{Materials and Methods}

\subsection{Study design, preregistration, and open materials}

This preregistered comparative methodological study was embedded in a scoping review of gaze semantics in human social interaction \citep{figalova_bockler_huestegge_2026}. The methodological study was preregistered on the Open Science Framework on February 24, 2026, after database searching and deduplication but before the preliminary title-only screen and all title-and-abstract screening procedures evaluated here.

The preregistration specified the principal human and LLM-based screening comparisons and analytical approach. Deviations from the preregistered plan and exploratory additions are documented in Supplementary Material S1. Additional open materials comprise detailed definitions of the outcome measures and statistical analyses (S2), the screening manual and version history (S3), input data (S4), complete LLM prompts (S5), screening outputs (S6), comprehensive analytical results (S7), the analysis script (S8), and the codebook (S9). These materials are publicly available on the Open Science Framework.

The complete search strategies and PRISMA-ScR reporting for the parent review are provided in the companion manuscript \citep{figalova_bockler_huestegge_2026}. The present study concerns the comparative performance of the human and LLM-based title-and-abstract screening workflows and therefore reports only the review procedures required to define the screening task, benchmark dataset, and reference outcomes.

\subsection{Benchmark dataset and analytical sets}
\label{dataset_creation}

Searches for the parent review were conducted in PubMed, PsycINFO, Web of Science Core Collection, and OSF Preprints on February 16, 2026, followed by a supplementary Web of Science search on February 23, 2026. After deduplication, the search corpus contained 5,291 records. The benchmark was constructed by the authors from bibliographic records retrieved for the parent scoping review and was not a pre-existing curated machine-learning or classification dataset.

The review lead (NF) conducted a conservative preliminary title-only screen to remove records whose titles clearly indicated that they did not meet the eligibility criteria. Records were retained whenever eligibility could not be determined reliably from the title alone. This preliminary screen removed 4,160 records and was used only to construct the benchmark dataset; it was not evaluated as a screening workflow in the present study.

The resulting benchmark comprised 1,131 records and is provided in Supplementary Material S4. The single-reviewer workflow, distributed-team workflow, and all seven LLM runs independently evaluated the same 1,131 records at the title-and-abstract level. Records identified subsequently through backward citation searching were considered in the parent review but were not added to the benchmark. All performance estimates therefore concern the 1,131 records remaining after the preliminary title-only screen, whose accuracy was not evaluated here.

Of the 1,131 benchmark records, 922 were selected for full-text retrieval and 209 were not advanced. Full-text reports were successfully obtained and assessed for 859 records; 63 could not be retrieved. Among the 859 full-text-assessed records, 316 were judged eligible and 543 ineligible.

Three analytical sets were therefore distinguished:

\begin{itemize}
\item the complete benchmark dataset ($n=1{,}131$), used to quantify retained workload, agreement, run-to-run consistency, and record-level decision overlap;
\item the full-text-assessed set ($n=859$), used for conditional reference-based classification measures; and
\item the verified eligible set ($n=316$), used to estimate operational recall and missed eligible records.
\end{itemize}

The relationship between outcome measures and analytical sets is summarised in Table~\ref{tab:outcome_measures_overview}.

\subsection{Data preprocessing}

The benchmark dataset was derived from bibliographic records retrieved for the parent scoping review. Database outputs were combined and deduplicated before the present study was preregistered. Following the conservative preliminary title-only screen described above, the remaining 1,131 records were assigned stable identifiers. For LLM screening, input files contained only the record identifier, title, and abstract. Records were divided into files or batches where required by the respective processing configuration, without substantive modification of the title or abstract text. No author-controlled tokenisation, embedding generation, feature engineering, model training, or fine-tuning was performed.

After screening, LLM outputs were checked for structural problems such as missing or duplicated identifiers, missing decisions, and malformed rows. Outputs were linked by stable record identifier and formatting was standardised where necessary. No LLM screening decision was manually changed on substantive grounds. For the primary analyses, \emph{Include} and \emph{Unclear} were subsequently mapped to \emph{retained}, whereas \emph{Exclude} and \emph{Exclude--citation seed} were mapped to \emph{not retained}.

\subsection{Screening task and decision structure}

An initial screening manual was developed before human calibration to operationalise the parent review's eligibility criteria and provide a common decision framework for the human and LLM-based workflows. The manual was refined following human calibration and finalised before formal title-and-abstract screening began. The complete manual and its version history are provided in Supplementary Material S3.

Records were considered potentially eligible when gaze or a related eye behaviour was substantively examined as carrying, expressing, signalling, cueing, shaping, or supporting the interpretation of meaning within a human social context. This included social interpretations of direct or averted gaze, eye contact, mutual gaze, gaze shifts, and gaze avoidance.

Records were excluded when their primary focus concerned developmental or clinical populations, psychiatric classification, non-social or low-level visual attention, methods-only gaze research, or technology evaluation without a central focus on the human interpretation of gaze meaning. Studies involving robots, avatars, or virtual agents remained eligible when they examined human social interpretations of gaze.

Each record received one of four mutually exclusive title-and-abstract decisions:

\begin{enumerate}
\item \textit{Include};
\item \textit{Unclear};
\item \textit{Exclude}; or
\item \textit{Exclude--citation seed}.
\end{enumerate}

\emph{Unclear} was assigned when the information available in the title and abstract was insufficient for a confident eligibility decision. \emph{Exclude--citation seed} identified an otherwise ineligible record that appeared potentially useful for supplementary backward citation searching.

For the primary quantitative analyses, \emph{Include} and \emph{Unclear} were mapped to \emph{retained}, whereas \emph{Exclude} and \emph{Exclude--citation seed} were mapped to \emph{not retained}. This binary mapping represented the operational decision of whether a record would proceed toward full-text assessment. Human reviewers and LLMs were also instructed to assign one primary exclusion reason to records classified as \emph{Exclude}. Exclusion reasons were collected for transparency and descriptive inspection but were not included in the primary quantitative comparison.

\subsection{Human screening workflows}

\subsubsection{Single-reviewer workflow}

The single-reviewer workflow was conducted by the review lead, NF, who screened all 1,131 benchmark records using the final screening manual and assigned one of the four predefined decisions to each record. Screening was conducted between March 2 and March 16, 2026.

\subsubsection{Distributed-team workflow}

The distributed-team workflow involved four psychology student assistants enrolled in a master's-level psychology programme. Each assistant screened a non-overlapping subset of the benchmark, and their decisions were subsequently combined to constitute one operational team workflow.

Before formal screening, NF and all four assistants independently assessed a calibration set of 28 records. The decisions were discussed in a joint meeting to identify sources of disagreement and establish a shared interpretation of the eligibility criteria. The screening manual was then refined and finalised. No second calibration round was conducted.

The calibration decisions themselves were not carried forward as final screening decisions. The 28 records remained in their original positions in the benchmark and were reassessed during formal screening by NF and by the assistant responsible for the corresponding file. No additional overlap subset, ongoing double-screening, or formal adjudication procedure was implemented during formal distributed-team screening.

The benchmark was divided sequentially, in the original Zotero-export order, into 12 non-overlapping Excel files: 11 files containing 100 records and one containing 31 records. Neither records nor files were randomised. Files were placed in a shared cloud folder and selected by assistants on a first-come-first-served basis; allocation was tracked separately to prevent overlap. The four assistants screened 300, 331, 200, and 300 records, respectively. Because assistants assessed different non-random subsets, assistant-level differences could reflect both reviewer behaviour and file composition and were therefore interpreted descriptively rather than as direct comparisons of reviewer performance.

The assistants were instructed not to use LLMs, automated summaries, AI-assisted classification tools, or other generative-AI applications. Distributed-team screening was conducted between March 2 and March 13, 2026. During formal screening, the assistants did not have access to NF's decisions, and NF did not have access to the assistants' decisions. All human title-and-abstract screening was completed before the LLM-based runs began.

Each assistant recorded their screening time. Aggregate distributed-team time represented summed person-time across assistants rather than elapsed calendar time.

\subsection{LLM-based screening workflows}

\subsubsection{Models and run design}

The LLM-based procedures were conducted through the standard hosted web interfaces of ChatGPT 5.4 Thinking, Gemini 3 Thinking, and Gemini 3.1 Pro. Model names are reported exactly as displayed in the respective interfaces at the time of data collection. The providers did not expose the hardware, GPU configuration, inference software stack, model weights, or complete backend version used for individual runs; these infrastructure details were therefore unavailable to the authors.

Seven complete LLM runs were conducted between March 17 and March 24, 2026. Each run constituted one complete pass through all 1,131 benchmark records.
The implemented runs formed a purposive, non-factorial set of model--configuration combinations. General-purpose reasoning-oriented LLMs available through standard hosted web interfaces were selected because the study aimed to evaluate screening procedures that could be implemented directly by review teams without review-specific model training, fine-tuning, or specialised computing infrastructure. Models from two provider ecosystems were included to avoid restricting the evaluation to a single hosted system. Processing configurations were selected to compare practically distinct ways of submitting the same screening task and to examine configuration effects and repeat-run consistency, rather than to estimate independent causal effects of model and processing configuration. Comparisons are therefore descriptive and refer to complete implemented workflows. The design was not intended to estimate independent causal effects of model identity and processing configuration. Accordingly, comparisons refer to complete implemented workflows rather than isolated model effects.

The seven runs and their analytical roles are summarised in Table~\ref{tab:llm_runs}. These included file-batch and all-at-once configurations for ChatGPT 5.4 Thinking and Gemini 3 Thinking, a nominally identical repeat of the initial ChatGPT 5.4 file-batch procedure, a Gemini 3.1 Pro file-batch run, and an interactive ChatGPT 5.4 configuration processing records sequentially in groups of 10.

Default interface settings were used. Memory was disabled, and user-adjustable sampling parameters were not modified. Input files contained only a stable record identifier, title, and abstract. Where records were divided across files, each file was processed in a separate conversation.

\subsubsection{Development of the LLM instructions}

The final human screening manual was translated into a structured LLM prompt specifying the eligibility criteria, decision categories, uncertainty rule, exclusion rules, and required output structure. The prompt was developed iteratively with assistance from ChatGPT 5.4 Thinking and was reviewed and refined by NF to ensure alignment with the human screening manual.

The prompt was tested procedurally using the same 28 records used for human calibration. This testing was limited to execution-related properties, including whether all requested records were processed, identifiers were preserved, required variables were returned, and the requested output structure was followed. Human screening decisions and full-text eligibility outcomes were not supplied to the models during prompt development or testing.

The 28 records remained in the benchmark and were reassessed in every formal LLM run. The calibration set was not used to select among prompt variants on the basis of agreement with human decisions, recall, or subsequent full-text eligibility. No systematic prompt-sensitivity analysis was conducted. Because ChatGPT 5.4 Thinking, one of the evaluated models, assisted with prompt development, the resulting estimates are interpreted as evaluations of the complete implemented workflows rather than model-independent estimates of screening capability. Complete prompts are archived in Supplementary Material S5.

\subsubsection{Processing configurations}

\paragraph{Approximately 100-record file batches.}

The benchmark was divided into the same 12 files used for distributed-team screening: 11 files containing 100 records and one containing 31 records. Each file was processed in a separate conversation. This configuration provided smaller independent outputs and allowed record coverage and output completeness to be inspected at the file level. The substantive screening criteria, decision categories, uncertainty rule, and requested output variables were held constant within the corresponding model-specific comparisons.

\paragraph{All records at once.}

One Excel file containing all 1,131 benchmark records was uploaded in a single conversation. The model was instructed to assess every record from its title and abstract and return a structured output containing the record identifier, title, abstract, screening decision, and, for excluded records, one predefined exclusion reason.

\paragraph{Interactive sequential groups of 10 records.}

The interactive configuration used ChatGPT 5.4 Thinking with the same screening manual, eligibility criteria, decision categories, and uncertainty rule. The model was instructed to process exactly 10 records per response, provide a decision and justification for each record, stop after each group, and wait for the instruction ``next.''

The prompt and screening manual (Supplementary Materials S5 and S3) were uploaded at the beginning of each conversation, followed by one approximately 100-record input file. NF entered only ``next'' between successive groups and provided no substantive feedback or correction. Once all records in a file had been assessed, the model was asked to generate an Excel file reproducing the decisions displayed in the conversation. Spreadsheet generation was treated as transcription of the preceding screening output rather than as an independent classification step.

\subsubsection{Output handling, integrity checks, and safeguards}

Stable record identifiers were preserved throughout the input, model-output, merging, and analysis pipeline. During import and merging, outputs were inspected for apparent problems in record coverage, duplicate or missing identifiers, missing decisions, and malformed rows. These integrity checks were used to identify obvious structural failures but were not conducted under a prespecified or fully systematic validation protocol. In the interactive sequential run, the final generated spreadsheets were not exhaustively compared record by record with every decision displayed earlier in the conversations. No unresolved structural problem was apparent in the final merged dataset.

Human title-and-abstract decisions, advancement status, full-text eligibility outcomes, and eventual inclusion in the parent review were absent from the files and prompts supplied to the models. LLM outputs were saved before they were linked to human decisions or full-text outcomes. Manual post-processing was restricted to identifier-based merging and formatting standardisation; no LLM screening decision was changed manually on substantive grounds.

Additional safeguards were used to reduce outcome leakage, record loss, and post hoc intervention. Memory was disabled; file-batch inputs were processed in separate conversations; stable identifiers were retained throughout; and the instructions explicitly required ambiguous or insufficiently described records to receive an \emph{Unclear} decision rather than a confident exclusion. These procedures were intended to reduce the risk of overconfident exclusion and contamination by study-specific reference outcomes.

Because the evaluated systems were proprietary hosted services, their training corpora were unavailable and possible prior exposure to the underlying publications or bibliographic records could not be assessed. The investigators also had no control over undocumented model updates or other server-side changes. The study therefore does not constitute external validation of the underlying models and cannot isolate model-specific effects from all properties of the hosted interfaces. Its inferential target is the set of model--prompt--processing-configuration workflows implemented under the conditions described here.

\subsection{Advancement to full-text assessment and reference outcomes}

Before the corresponding workflows were run and before their outputs were inspected, the parent review prospectively designated four advancement workflows:

\begin{itemize}
\item the single-reviewer workflow;
\item the distributed-team workflow;
\item Gemini 3 Thinking using approximately 100-record file batches; and
\item the initial ChatGPT 5.4 Thinking approximately 100-record file-batch run.
\end{itemize}

A record was selected for full-text retrieval when at least one of these four workflows assigned \emph{Include} or \emph{Unclear}. Thus, advancement followed a liberal union rule and did not require agreement, majority voting, or consensus adjudication. Records assigned \emph{Exclude} or \emph{Exclude--citation seed} by all four advancement workflows were not selected for retrieval.

The file-batch LLM configurations were prospectively selected as advancement workflows because their smaller, separate outputs were expected to permit more transparent inspection of record coverage and output completeness. The all-at-once configurations were designated as comparative runs. Their outputs were subsequently found to be structurally complete enough for inclusion in the methodological comparison, but they did not contribute to retrieval decisions.

The retrieval set was fixed before the nominally identical ChatGPT file-batch repeat run, the Gemini 3.1 Pro run, and the interactive sequential run were conducted. Decisions from these later comparative runs therefore did not retrospectively alter which records underwent full-text retrieval.

NF alone assessed the 859 retrieved full-text reports against the original eligibility criteria. Workflow- and run-specific title-and-abstract decisions were not displayed during full-text assessment. Because NF had previously completed the single-reviewer title-and-abstract screen, however, full-text assessment should not be interpreted as fully blinded to all prior screening experience. The operational reference standard therefore relied on a single full-text assessor. Full-text decisions and primary exclusion reasons were recorded using a standardised decision form and exclusion-reason scheme.

Final eligibility was verified only for records selected by the union of the four advancement workflows and successfully retrieved. Eligible records missed by all four advancement workflows could therefore remain unidentified among the 209 non-advanced records, while the eligibility of the 63 unretrieved records remained unknown. Reference-based measures consequently quantify performance within the implemented review pathway rather than accuracy against a fully and independently verified benchmark. Comparisons involving later or comparative LLM runs are correspondingly not fully symmetric.

\subsection{Outcome measures and statistical analysis}

For the analyses, \emph{screening output} denotes the decisions produced by one complete human workflow or LLM run. The nine screening outputs comprised two human workflows and seven LLM runs. Table~\ref{tab:outcome_measures_overview} summarises the outcome measures, their interpretation, and the records contributing to each analysis. Detailed definitions, formulas, and additional interpretive limitations are provided in Supplementary Material S2.

The primary screening trade-off was evaluated jointly using retained workload and operational recall. Retained workload was the number and proportion of the 1,131 benchmark records classified as \emph{Include} or \emph{Unclear} and therefore retained for possible full-text assessment. Operational recall quantified the proportion of the 316 verified eligible records retained by each workflow. Because final eligibility was not verified across the complete benchmark, this measure is an operational estimate rather than complete-benchmark recall.

Conditional reference-based classification performance was evaluated among the 859 records with verified full-text outcomes. These analyses included precision, specificity, and $F_1$ alongside recall. Their interpretation is conditional on full-text verification and therefore does not extend to the 209 non-advanced or 63 unretrieved records with unknown final eligibility.

Agreement was evaluated independently of final eligibility across the complete benchmark. All 36 pairwise comparisons among the nine screening outputs were analysed. Pairwise retained/not-retained agreement was summarised using overall agreement, Cohen's $\kappa$, Gwet's AC1, and positive and negative agreement. AC1 was included alongside $\kappa$ because $\kappa$ can be strongly influenced by category prevalence and marginal decision distributions \citep{byrt1993bias,gwet2008computing}.

Run-to-run consistency was examined specifically for the two ChatGPT 5.4 file-batch runs conducted under nominally identical conditions. In addition to aggregate performance, the analysis examined whether the same individual records were retained across runs and whether discordant decisions involved verified eligible records.

Record-level analyses further examined the overlap of missed eligible records, records uniquely recovered by individual outputs, and complementarity between workflows. Pairwise liberal-union analyses treated a record as retained when either member of a pair retained it, allowing the gain in recovered eligible records to be considered jointly with the corresponding increase in retained workload.

Ninety-five per cent confidence intervals were estimated using 10,000 resamples. For operational recall, precision, specificity, and $F_1$, resampling was stratified by full-text eligibility status within the full-text-assessed set. For agreement measures, confidence intervals were estimated through multinomial resampling of the corresponding observed $2 \times 2$ table. Percentile intervals were obtained from the resulting empirical distributions.

Because the complete constructed benchmark was analysed, these confidence intervals do not quantify uncertainty about the fixed observed values within this corpus. Instead, they describe resampling variability when the observed records are treated as an empirical distribution and may support cautious generalisation to comparable records. The intervals do not incorporate uncertainty about the eligibility of non-advanced or unretrieved records and do not account for potential clustering by assistant, source file, conversation, or multiple reports from the same underlying study.

No paired hypothesis tests or between-workflow contrast intervals were calculated. Consequently, overlap or non-overlap between confidence intervals for separate workflows was not interpreted as statistical evidence of a difference between them.

As a sensitivity analysis, the principal retained-workload, operational-recall, missed-record, conditional-classification, and agreement analyses were repeated after excluding the 28 records used for human calibration and procedural prompt testing, with all relevant denominators recalculated.

No a priori power calculation was conducted because the complete available benchmark was analysed and the study was designed as a descriptive methodological comparison rather than a confirmatory hypothesis test.

\subsection{Software and computational reproducibility}

Analyses were conducted in R version 4.6.1 (2026-06-24 ucrt) using RStudio 2026.06.0 on Windows 11. Packages included \emph{tidyverse} version 2.0.0, \emph{knitr} version 1.51, and \emph{ggrepel} version 0.9.8. The random seed used for resampling was 20260703.

The analysis code, benchmark input data, LLM prompts, screening outputs, detailed statistical definitions, and comprehensive analytical results are archived in the accompanying open materials (Supplementary Materials S2 and S4--S8), allowing the reported comparisons to be reconstructed from the archived screening outputs. Re-execution of the proprietary hosted LLM workflows cannot be guaranteed to reproduce the original outputs because the underlying models and web interfaces are externally controlled and may change over time.

\section{Results}

Results are organised according to the five research questions. Exploratory analyses of pairwise workflow combinations, descriptive measures of screening effort and procedural burden, and the sensitivity analysis excluding calibration and prompt-testing records are reported subsequently. Detailed results, including complete pairwise comparisons, are provided in the Supplementary Materials.

\subsection{RQ1: Recovery of verified eligible records and retained workload}
\label{ResultsRQ1}

\begin{figure}[htbp]
\centering
\includegraphics[width=0.99\linewidth]{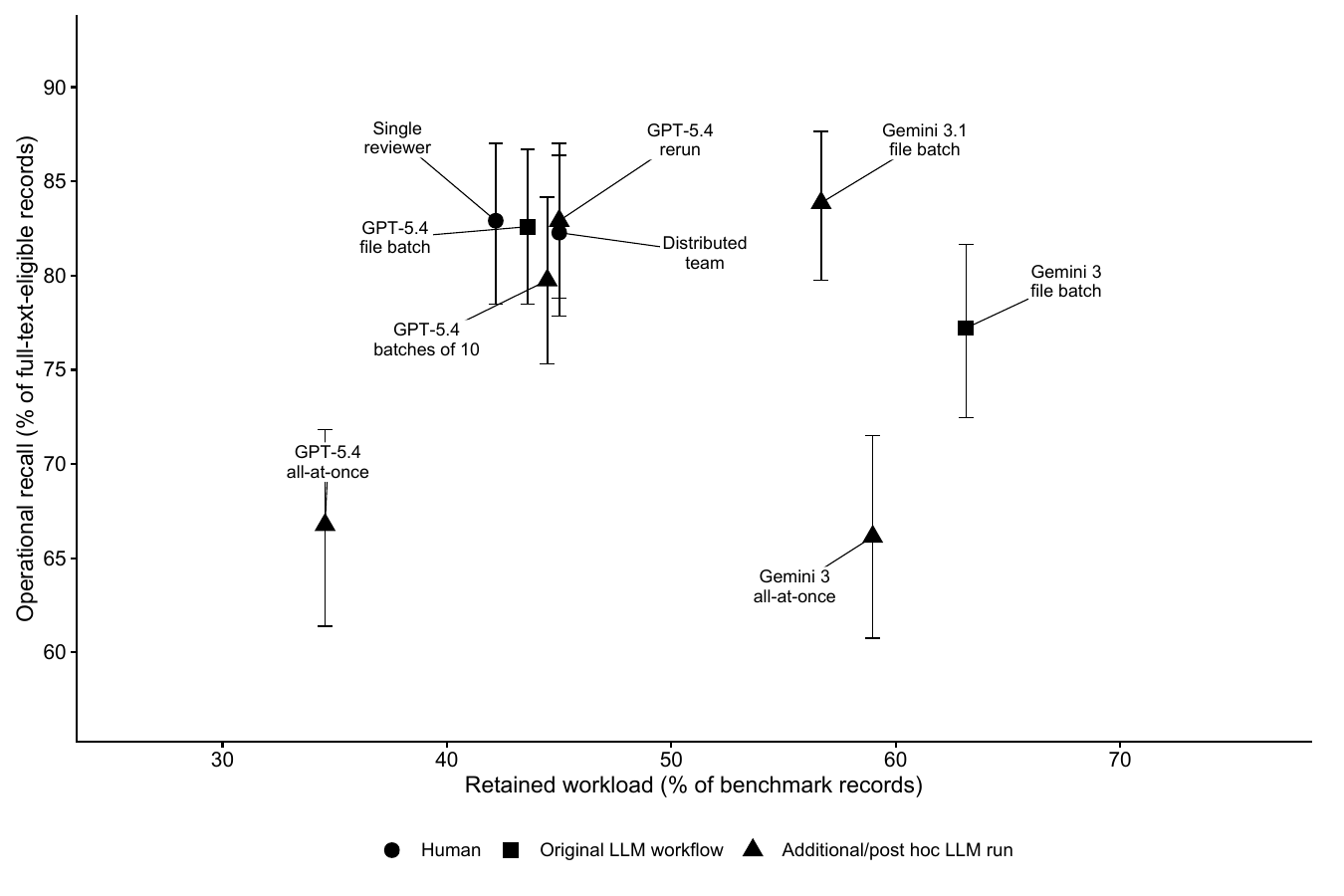}
\caption{Operational recall as a function of retained screening workload across human and LLM-based screening workflows.}
\label{fig:recall_workload_tradeoff}
\end{figure}

The principal comparison considered operational recall jointly with retained workload (Table~\ref{tab:workload_recall}; Figure~\ref{fig:recall_workload_tradeoff}). No individual screening output recovered all 316 verified eligible records, and greater retention of benchmark records did not consistently correspond to greater recovery.

The two human workflows and the two GPT-5.4 file-batch runs occupied a closely situated region of the workload--recall space. They retained 42.2--45.0\% of the 1,131-record benchmark while recovering 82.3--82.9\% of the verified eligible records. Within this group, the single reviewer retained the fewest benchmark records. The interactive GPT-5.4 batches-of-10 run had a similar retained workload (44.5\%) but lower operational recall (79.7\%).

Gemini 3.1 file batches achieved the highest operational-recall point estimate (83.9\%) but retained 56.7\% of the benchmark. Gemini 3 file batches retained the largest proportion of records overall (63.1\%) while recovering 77.2\% of the verified eligible set. Thus, the most liberal workflow did not achieve the highest recovery.

Within both model families for which file-batch and all-at-once configurations were evaluated, the file-batch runs recovered more verified eligible records. GPT-5.4 file batches achieved 82.6\% operational recall compared with 66.8\% for GPT-5.4 all at once. Gemini 3 file batches achieved 77.2\% compared with 66.1\% for Gemini 3 all at once. The associated workload patterns differed, however: GPT-5.4 all at once produced the lowest retained workload of all outputs (34.6\%), whereas Gemini 3 all at once retained 59.0\% of the benchmark. These results indicate substantial differences between processing configurations but, given the non-factorial design, do not isolate configuration as their causal source.

\subsection{RQ2: Agreement between screening workflows}
\label{ResultsRQ2}

Binary retained/not-retained agreement varied considerably across workflow pairs (Table~\ref{tab:main_agreement_summary}). Agreement between the single reviewer and distributed team was 72.8\% ($\kappa=0.447$; AC1 = 0.464). The GPT-5.4 file-batch workflow showed a similar level of agreement with the two human workflows: 71.2\% with the single reviewer and 74.0\% with the distributed team. By contrast, Gemini 3 file batches agreed with the single reviewer and distributed team on 55.0\% and 55.7\% of records, respectively.

Runs using the same displayed model but different processing configurations also produced different decisions. Agreement between GPT-5.4 file batches and GPT-5.4 all at once was 82.5\%, whereas agreement between the two Gemini 3 configurations was 62.4\%. Thus, sharing the same displayed model did not imply closely matching screening outputs across processing configurations.

The highest agreement among the selected comparisons occurred between the two GPT-5.4 file-batch runs conducted under nominally identical conditions (91.7\%). Their remaining record-level disagreement is examined under RQ4.

Agreement did not map directly onto recovery of eligible records. Workflows with similar overall agreement could differ in which records they retained and in the consequences of their discordant decisions, supporting separate consideration of agreement and reference-based performance.

\subsection{RQ3: Conditional classification performance among full-text-assessed records}
\label{ResultsRQ3}

Reference-based classification measures were calculated within the 859 records for which full-text eligibility was verified, comprising 316 eligible and 543 ineligible records (Table~\ref{tab:conditional_classification_results}). Because all 316 verified eligible records belonged to this set, recall was numerically identical to the operational recall reported under RQ1.

The single reviewer achieved the highest conditional $F_1$ score (0.693). The initial GPT-5.4 file-batch run ($F_1=0.670$), its rerun ($F_1=0.667$), and the distributed-team workflow ($F_1=0.663$) produced broadly similar conditional profiles. The interactive GPT-5.4 batches-of-10 run was slightly lower ($F_1=0.653$).

GPT-5.4 all at once generated the fewest false positives and achieved the highest conditional specificity (0.718), but its lower recall resulted in an $F_1$ score of 0.621. Conversely, Gemini 3.1 file batches achieved the highest operational recall but retained more full-text-assessed ineligible records, yielding lower precision and specificity.

The two Gemini 3 configurations had the lowest conditional $F_1$ scores overall. Gemini 3 file batches showed the lowest precision and specificity, whereas Gemini 3 all at once produced the lowest $F_1$ score.

Accordingly, no single metric identified a uniformly superior workflow. Higher specificity could coincide with substantial loss of verified eligible records, while higher recall could require retaining substantially more ineligible records. These estimates are conditional on the 859 full-text-assessed records and should not be interpreted as complete-benchmark classification performance.

\subsection{RQ4: Run-to-run consistency of the repeated GPT-5.4 file-batch workflow}
\label{ResultsRQ4}

The two GPT-5.4 file-batch runs conducted under nominally identical conditions produced closely similar aggregate performance but differed at the record level. Across the 1,131 benchmark records, the runs agreed on 1,037 records (91.7\%) and disagreed on 94 (8.3\%). They jointly retained 454 records and jointly did not retain 583; among the discordant records, 39 were retained only by the first run and 55 only by the rerun.

Aggregate retained workload, operational recall, precision, specificity, and $F_1$ were closely aligned between the two runs (Tables~\ref{tab:workload_recall} and~\ref{tab:conditional_classification_results}). Their differences were therefore more apparent at the level of individual records than in the summary measures.

Among the 316 verified eligible records, 247 were retained by both runs, 14 only by the first run, and 15 only by the rerun; 40 were missed by both. Thus, 29 verified eligible records received discordant advancement decisions despite the nominally identical procedure. Combining the two runs under a liberal union rule would have recovered 276 verified eligible records (87.3\%) while retaining 548 benchmark records (48.5\%).

These results show that similar aggregate performance did not imply reproducible record-level classifications across repeated executions of the same hosted LLM workflow.

\subsection{RQ5: Missed verified eligible records and retrospective checks}
\label{ResultsRQ5}

\subsubsection{Overlap and unique recovery}

Missed eligible records were not distributed uniformly across screening outputs. Among the 316 verified eligible records, 84 (26.6\%) were retained by all nine outputs and 91 (28.8\%) were missed by only one. In total, 175 records (55.4\%) were therefore retained by at least eight of the nine outputs. At the opposite extreme, seven verified eligible records were recovered by only one output.

All seven of these uniquely recovered eligible records were retained by a human workflow: six only by the single reviewer and one only by the distributed team. No verified eligible record was uniquely recovered by an LLM output while being missed by both human workflows.

By design, no verified eligible record could have been missed by all nine outputs, because final eligibility could only be established for records advanced by at least one of the four original advancement workflows. The observed distribution therefore characterises disagreement within the verified eligible set and cannot quantify potentially eligible records missed by every advancement workflow.

\subsubsection{Post hoc review of records retained only by the interactive LLM run}

Of the complete benchmark, 209 records (18.5\%) were not retained by any of the four original advancement workflows and were consequently not selected for full-text retrieval. The later GPT-5.4 batches-of-10 run retained 13 of these records. Had this run contributed to the advancement rule, the retrieval set would therefore have increased from 922 to 935 records.

Post hoc reassessment of these 13 titles and abstracts found that 11 were clearly outside the review scope, primarily because they concerned clinical or diagnosis-defined samples, developmental research, human--computer interaction, or were not substantively focused on gaze. Two records could not be confidently excluded from the title and abstract alone and, retrospectively, would have warranted full-text retrieval.

Because this assessment was conducted after the retrieval set and analytical denominators had been fixed, it did not alter the operational recall or conditional classification estimates. It nevertheless provides direct evidence that the set of non-advanced records could contain potentially relevant records whose final eligibility remained unverified.

\subsection{Exploratory analysis: pairwise combined recovery}
\label{ResultsExploratory}

Combining screening outputs under a liberal union rule consistently increased recovery relative to the constituent workflows but also increased retained workload. The magnitude of this trade-off depended strongly on the specific pairing.

The two human workflows together recovered 310 of 316 verified eligible records (98.1\%) while retaining 647 benchmark records (57.2\%). Of the recovered eligible records, 212 were retained by both workflows, 50 only by the single reviewer, and 48 only by the distributed team; six verified eligible records were missed by both.

Combining the single reviewer with the initial GPT-5.4 file-batch workflow recovered 307 verified eligible records (97.2\%) while retaining 648 benchmark records (57.3\%), a nearly identical workload to the human--human pair but with three fewer eligible records recovered. Combining the distributed team with GPT-5.4 file batches retained the same number of benchmark records but recovered 296 eligible records (93.7\%).

The single reviewer combined with Gemini 3.1 file batches recovered 310 verified eligible records, matching the human--human pair, but retained 762 benchmark records (67.4\%). This represented 115 additional retained records for the same recovery. Pairing either human workflow with Gemini 3 file batches resulted in still greater retained workloads.

Among selected LLM-only combinations, Gemini 3 file batches combined with GPT-5.4 file batches recovered 295 verified eligible records while retaining 820 benchmark records. Combining the two GPT-5.4 file-batch runs recovered 276 eligible records while retaining 548 benchmark records.

Thus, complementarity was present across several human and LLM combinations, but additional recovery could not be considered independently of the extra full-text workload generated. Among the pairs recovering 310 verified eligible records, the human--human combination required the lowest retained workload.

\subsection{Screening effort and procedural burden}
\label{ResultsSupplementaryDescriptive}

\subsubsection{Human screening effort}

The single reviewer screened all 1,131 records in 1,176 minutes of logged active screening time, corresponding to 1.04 minutes per record or 57.7 records per hour (Table~\ref{tab:human_screening_time}). The distributed-team workflow required 1,628 minutes of summed person-time, corresponding to 1.44 minutes per record or 41.7 records per hour. Individual assistant screening rates ranged from 39.3 to 44.4 records per hour.

These estimates refer to formal screening only. The single-reviewer measure excluded breaks and recovery time, while the distributed-team estimate excluded approximately three hours devoted to the introductory meeting, independent calibration screening, and subsequent calibration discussion. Distributed-team time represents summed person-time and is therefore not equivalent to elapsed calendar time.

\subsubsection{LLM procedural burden}

LLM timing measures describe interface processing and output generation rather than continuous human labour and are therefore not directly comparable with manual screening time. Active user interaction was not logged prospectively and is reported only as an approximate retrospective estimate.

GPT-5.4 all at once required 46 minutes and 29 seconds of visible screening time and 5 minutes and 45 seconds for spreadsheet generation. Because this procedure involved a single conversation, active user interaction was estimated at approximately one to two minutes.

The initial GPT-5.4 file-batch run used 12 conversations and required 66 minutes of visible screening time plus 8 minutes and 36 seconds for spreadsheet generation. The nominally identical rerun required 52 minutes plus 7 minutes and 18 seconds, respectively. Active user interaction for each file-batch run was estimated at approximately one to two minutes per conversation, or approximately 12--24 minutes in total.

The interactive GPT-5.4 batches-of-10 run required 92 minutes and 54 seconds of visible screening time and approximately 49 minutes for spreadsheet generation. Because the model had to be prompted repeatedly to continue, active interaction was estimated at approximately five minutes per file. Combining batch outputs into the analytical dataset required approximately five additional minutes per run.

Comparable timing data were unavailable for the Gemini workflows. Although the recorded GPT-5.4 procedures required substantially less active human interaction than manual screening, the comparison remains descriptive because human and LLM timing measures represented different forms of effort and were not collected using a common timing protocol.

\subsection{Sensitivity analysis excluding calibration and prompt-testing records}
\label{ResultsSensitivity}

Excluding the 28 records used for human calibration and procedural LLM prompt testing left 1,103 benchmark records, including 306 verified eligible records. The principal findings were unchanged.

Across workflows, retained workload changed by between $-0.7$ and $0.2$ percentage points, operational recall by between $-0.003$ and $0.007$, precision by between $-0.003$ and $0.004$, specificity by between $-0.005$ and $0.007$, and $F_1$ by between $-0.003$ and $0.003$. Workflow ordering on the principal measures remained unchanged: Gemini 3.1 file batches retained the highest operational-recall point estimate, the single reviewer retained the highest conditional $F_1$ score, and GPT-5.4 all at once retained the highest specificity while continuing to show substantially lower recall than the file-batch workflows.

Agreement estimates were similarly stable. Agreement between the single reviewer and distributed team was 73.0\% after exclusion, compared with 72.8\% in the primary analysis. Agreement between the two GPT-5.4 file-batch runs remained 91.7\%, and agreement between the distributed team and the initial GPT-5.4 file-batch run was 74.1\%. Excluding the calibration and prompt-testing records therefore did not materially alter the principal performance or agreement patterns.
\section{Discussion}

\subsection{Principal findings}

This preregistered methodological study compared two human workflows and seven LLM-based screening runs in a conceptually complex title-and-abstract classification task. The results show that screening performance was a property of the implemented workflow rather than of the displayed model alone. Across the evaluated procedures, recovery of verified eligible records, retained workload, agreement, conditional classification performance, complementarity, and run-to-run consistency captured distinct aspects of performance, and no individual output recovered all 316 verified eligible records.

The two human workflows and the two GPT-5.4 file-batch runs occupied a similar region of the workload--recall space, retaining approximately 42--45\% of the benchmark while recovering approximately 82--83\% of the verified eligible records. Gemini 3.1 file batches achieved the highest operational-recall point estimate, but at substantially greater retained workload. In contrast, both all-at-once configurations recovered only about two thirds of the verified eligible records. Greater retention therefore did not translate consistently into greater recovery, and the configuration with the lowest retained workload also missed a substantial proportion of the verified eligible set.

Processing configuration was associated with marked differences within both model families for which file-batch and all-at-once procedures were compared. File-batch processing recovered more verified eligible records than all-at-once processing for both GPT-5.4 and Gemini 3, although the corresponding workload patterns differed. Because model, interface, date, interaction structure, and processing configuration were not varied independently, these comparisons do not establish a causal effect of batching. They nevertheless demonstrate that performance observed for one implementation of a model should not be assumed to generalise automatically to another implementation of the same displayed model.

Independent screening outputs were also partly complementary. A liberal union of the two human workflows recovered 98.1\% of the verified eligible records while retaining 57.2\% of the benchmark. Some human--LLM combinations approached or matched this level of recovery, but generally at greater retained workload. Repeating the GPT-5.4 file-batch workflow also increased combined recovery relative to either run alone, but the resulting operating point differed from that of the human--human combination. The practical value of combining outputs therefore depended on both the specific pairing and the relative cost assigned to missed evidence versus additional downstream assessment.

The repeated GPT-5.4 file-batch runs provide a further important result. Their aggregate workload, recall, and conditional classification measures were closely aligned, and they agreed on 91.7\% of benchmark records. Nevertheless, they disagreed on 94 individual records, including 29 verified eligible records retained by only one run. Aggregate similarity therefore concealed meaningful item-level instability. Evaluations of LLM-based classifiers should consequently consider not only summary performance but also whether repeated executions identify the same individual cases.

\subsection{Implications for LLM workflow evaluation and deployment}

A central implication is that the model name is an insufficient description of an LLM-based decision system. The implemented workflow also comprises input preparation, prompt instructions, uncertainty handling, processing configuration, interaction structure, output generation, integrity checks, post-processing, and the rule by which classifications are translated into downstream actions. Differences in any of these components may alter the behaviour of the deployed system. This interpretation is consistent with evidence that prompt design, output format, decision thresholds, and batch size can affect LLM screening outcomes \citep{dennstadt2024title,syriani2024screening,fagerberg2026batch}.

Processing configuration should therefore be documented and evaluated as a substantive workflow characteristic rather than treated as an incidental implementation detail. In the present study, file batching also provided practical advantages for auditing record coverage and inspecting separate outputs. However, the observed differences cannot be attributed specifically to batch size. They could reflect context length, allocation of model attention across records, response-generation constraints, interface behaviour, or interactions among these factors. Likewise, the study does not establish superiority of one model family over another because model, interface, processing configuration, interaction structure, and run date were not manipulated factorially, and the common prompt was developed with assistance from ChatGPT 5.4 Thinking. The appropriate inferential target is therefore the complete model--prompt--configuration workflow.

The findings also illustrate why LLM performance should not be represented by a single accuracy-like measure. Operational recall quantified recovery of the verified eligible set, retained workload quantified the number of records that would proceed to later assessment, and agreement quantified similarity between classifiers without indicating whether their shared decisions preserved relevant evidence. Precision, specificity, and $F_1$ described classification only among records for which full-text outcomes were available.

These measures favoured different workflows. Gemini 3.1 file batches achieved the highest operational-recall point estimate but retained substantially more records. GPT-5.4 all at once retained the fewest records and achieved comparatively high conditional specificity, yet missed approximately one third of the verified eligible set. In a high-recall task with asymmetric error costs, an apparently efficient reduction in workload may therefore result from an undesirable increase in false-negative decisions. Conversely, high recall can be achieved inefficiently by advancing many records. Evaluation should consequently report recovery and missed cases jointly with the workload required to achieve them rather than optimise either quantity in isolation \citep{madeyski2025llm4screenlit,sanghera2025high}.

Agreement likewise requires careful interpretation. High overall agreement can be dominated by records on which both workflows agree not to retain, particularly when ineligible records are common. More importantly, the repeated-run analysis shows that even relatively high overall agreement may coexist with disagreement on cases that have disproportionate downstream consequences. For stochastic or otherwise nondeterministic LLM workflows, record-level consistency is therefore a separate property from aggregate predictive performance.

This distinction has broader relevance for LLM-based decision support. If individual classifications determine which cases receive further assessment, two runs with similar aggregate metrics are not operationally interchangeable when they select meaningfully different cases. Reporting only average recall, precision, or agreement may obscure this instability. Repeated-run evaluations should therefore report the number and characteristics of discordant cases, including high-consequence cases classified differently across runs, rather than relying solely on differences between aggregate estimates.

The complementarity analyses further show that combining classifiers changes the operating point rather than simply ``improving accuracy.'' Liberal union rules increased recovery but also increased the number of records requiring downstream assessment. The two human workflows were particularly complementary: together they recovered substantially more eligible records than either alone while producing a more favourable workload--recovery balance than the human--LLM combination that reached the same recovery. LLM outputs nevertheless contributed additional decisions not shared with individual human workflows, consistent with previous evidence that human--LLM and LLM--LLM combinations can increase sensitivity \citep{sanghera2025high}.

Alternative combination rules would optimise different objectives. For example, systems that accept automated decisions only when multiple LLMs agree and refer discordant cases to humans place greater emphasis on reducing human workload while retaining human oversight of uncertain cases \citep{hilkenmeier2026full}. The appropriate architecture therefore depends on the relative costs of false negatives, false positives, and human review rather than on the assumption that one combination strategy is universally preferable.

Among the 316 verified eligible records, all seven records recovered by only one of the nine outputs were uniquely retained by a human workflow. This result should not be interpreted as evidence that LLMs cannot identify eligible records missed by humans because final eligibility was not established for every non-advanced record. The later interactive GPT-5.4 run retained 13 records that had not been advanced by any of the four original workflows. Post hoc title-and-abstract reassessment suggested that 11 were clearly outside scope, whereas two could not be confidently excluded without full-text assessment. Their final eligibility remains unknown. This illustrates an important consequence of partially verified reference standards: records missed by all workflows contributing to verification can remain invisible to standard recall calculations.

For deployment, the present results do not support treating an unvalidated LLM run as an autonomous exclusion component in a conceptually complex, high-recall screening task. A more defensible role is as one component of an auditable decision-support system, for example for supplementary screening, prioritisation, or identification of uncertain cases. Such systems require stable identifiers, structured machine-readable outputs, explicit uncertainty rules, checks for missing and duplicate records, transparent post-processing and merging procedures, predefined decision rules, and validation within the intended application context.

Reporting should correspondingly describe enough of the complete pipeline to make the evaluated system interpretable: model and interface version, prompts, input preparation, processing configuration, interaction structure, human involvement, output handling, advancement or decision rules, recovery, missed cases, downstream workload, agreement, and repeated-run consistency \citep{susnjak2023prisma,shailendra2025lprisma,holst2025prismatraice,gallifant2025tripodllm,luo2025gamer}. Reporting a model name and prompt alone may be insufficient when other workflow components materially affect the resulting decisions.

No universally optimal human--LLM arrangement is therefore implied by these findings. In applications where false negatives are particularly costly, a liberal combination rule may justify additional downstream workload. Where human resources are more constrained, prioritisation or selective human review of uncertain or discordant cases may be more appropriate. The optimal operating point depends on the application, error costs, conceptual complexity, available resources, reviewer expertise, and tolerance for missed cases.

\subsection{Strengths and limitations}

A principal strength of this study is that it was preregistered and embedded in an actual PRISMA-ScR-aligned review rather than evaluated solely on a retrospectively assembled or artificial benchmark. All nine screening outputs classified the same 1,131 records using the same substantive eligibility criteria and a common operational retained/not-retained rule. The comparison included a solo expert reviewer, a distributed trained human team, multiple practically implementable LLM workflows, different processing configurations, and a nominally repeated LLM workflow.

A further strength is the multidimensional evaluation framework. Recovery, retained workload, agreement, conditional classification performance, pairwise complementarity, and run-to-run consistency were analysed separately rather than collapsed into one performance score. This distinction made it possible to identify cases in which superficially favourable performance on one measure corresponded to an unfavourable trade-off on another. The sensitivity analysis excluding the 28 calibration and prompt-testing records did not materially alter the principal findings.

The study also used a conservative treatment of the reference outcomes. Records without verified full-text outcomes were not automatically classified as true negatives. This avoided producing apparently complete benchmark metrics by assuming that every record excluded before full-text assessment had been excluded correctly.

The corresponding limitation is that the operational reference set was conditional on the parent review's advancement and retrieval procedures. Final eligibility was not independently established for all 1,131 benchmark records. Potentially eligible records may therefore have remained among the 209 non-advanced records or the 63 records whose full texts could not be retrieved. Operational recall consequently represents recovery of the known verified eligible set rather than complete-benchmark sensitivity, while precision, specificity, and $F_1$ are conditional on the full-text-assessed subset.

The four-workflow advancement union was selected prospectively before the relevant outputs were inspected, but this specific union was not specified in the preregistration. In addition, those four advancement workflows contributed to construction of the verified eligible set, whereas the later comparative LLM runs did not. Comparisons between advancement and comparative workflows are therefore partly asymmetric.

Full-text eligibility was assessed by the review lead rather than by multiple independent assessors. Although workflow-specific title-and-abstract decisions were not displayed during full-text assessment, the review lead had previously completed the single-reviewer title-and-abstract screen. The verified eligible set should therefore be interpreted as an operational reference standard rather than an independent consensus gold standard.

The human results are likewise specific to the implemented arrangements. The review lead had substantially greater familiarity with the conceptual scope, whereas the distributed workflow comprised four calibrated master's-level psychology assistants who screened non-overlapping subsets without ongoing double-screening or formal adjudication. Differences among assistants could also be confounded with the non-random record subsets they screened. Other training, expertise, overlap, or consensus procedures could produce different human performance.

The LLM workflows were executed through proprietary hosted web interfaces. This increased ecological validity for researchers using readily available systems but reduced experimental control and strict reproducibility. Underlying model identifiers, weights, training corpora, server-side settings, and undocumented model updates were unavailable. The repeated GPT-5.4 workflow should consequently be understood as a nominal repetition of the observable procedure rather than a strict computational replication.

The non-factorial design further limits causal interpretation. Model identity, processing configuration, interaction structure, interface, and date were not independently manipulated. The study can therefore establish that the evaluated workflows behaved differently, but not determine which individual component caused those differences. Similarly, because the common LLM prompt was developed with assistance from ChatGPT 5.4 Thinking, performance should not be interpreted as arising from a model-neutral prompt-development process.

Output-integrity validation was also imperfect. Stable identifiers were maintained and outputs were inspected for missing records, duplicated identifiers, missing classifications, and malformed rows, but these checks were not conducted under a fully prespecified validation protocol. For the interactive batches-of-10 workflow, the final generated spreadsheets were not exhaustively compared record by record with every classification previously displayed in the conversations. Although no unresolved structural problem was apparent in the final analytical dataset, future evaluations should incorporate automated and prespecified completeness checks and verify exported outputs directly against the classifications generated during model interaction.

Procedural burden was not measured using a common protocol across human and LLM workflows. Human measures represented active screening or summed person-time, whereas LLM measurements included visible interface processing, output generation, waiting, and subsequent data handling. Active LLM interaction was estimated retrospectively, and comparable timing was unavailable for the Gemini workflows. The study therefore supports only a descriptive comparison of procedural burden and cannot establish standardised time or economic efficiency.

Generalisability is limited by the use of one interdisciplinary and conceptually complex scoping review. The benchmark was also constructed after a preliminary title-only screen removed 4,160 of the 5,291 deduplicated records. The resulting 1,131-record benchmark was therefore enriched for records that could not be confidently excluded from their titles alone. This may have affected difficulty, class prevalence, retained workload, and agreement and may also have advantaged the review lead through previous exposure to the complete search corpus.

The study did not systematically vary prompts, languages, eligibility contexts, or review datasets and did not include external validation in an independent corpus. Possible prior exposure of the proprietary models to the underlying publications or bibliographic records could not be assessed because their training data were unavailable. Human screening decisions, advancement status, full-text outcomes, and eventual inclusion in the parent review were, however, not supplied to the models.

The findings should therefore be interpreted as evidence about the specific workflows evaluated here rather than as general validation of the displayed models for literature screening or document classification. Future studies would benefit from independent datasets, prespecified factorial comparisons of workflow components, multiple prompt variants, systematic output-integrity checks, complete or independently sampled verification of non-advanced records, and repeated executions sufficient to characterise record-level variability.

Finally, the findings reflect models and hosted interfaces available in March 2026. Rapid changes in proprietary LLM systems limit the durability of model-specific performance estimates. This strengthens the case for evaluating and documenting the complete workflow actually deployed rather than relying on performance reported previously for a model carrying the same or a similar name.

\subsection{Conclusion}

In this conceptually complex screening task, the two GPT-5.4 file-batch runs produced workload--recall profiles close to those of the human workflows, whereas the all-at-once configurations recovered substantially fewer verified eligible records. No individual output recovered the complete verified eligible set, and greater recovery sometimes required substantially greater downstream workload. Recovery, workload, agreement, conditional classification performance, complementarity, and run-to-run consistency therefore represented distinct dimensions of screening performance.

The AI approach evaluated here was the prompt-based application of general-purpose proprietary LLMs rather than a review-specific model trained or fine-tuned for screening. This approach was selected because it represents an immediately implementable form of LLM-assisted screening that does not require labelled training data, local model deployment, or specialised machine-learning infrastructure. The findings consequently concern the performance of complete implemented screening workflows rather than a newly developed model architecture or the intrinsic capabilities of the displayed models.

The two human workflows were particularly complementary, jointly recovering 98.1\% of verified eligible records while retaining fewer records than the human--LLM combination that achieved the same recovery. LLM workflows also contributed complementary decisions, but their value depended on the workflow with which they were combined and the resulting workload. Moreover, similar aggregate results across nominally repeated LLM runs concealed meaningful differences in the individual records retained.

These findings support using LLMs as documented, auditable, and human-supervised components of evidence-synthesis workflows rather than as autonomous replacements for human screening judgement. For conceptually complex reviews, the relevant question is therefore not simply which model is used, but how the complete workflow is implemented, validated, combined with human judgement, assessed for record-level variation, and transparently reported. The present findings are specific to the evaluated review, models, prompts, interfaces, and processing configurations and should not be interpreted as external validation of the displayed models across evidence-synthesis contexts. Future evaluations should test such workflows prospectively across independent review topics and model versions, with more complete reference verification and systematic variation of prompts and processing configurations.


\section*{Competing interests}

The authors declare that they have no competing interests.

\section*{Author contributions}

\noindent\textbf{Nikol Figalová:}
Conceptualization, Methodology, Investigation, Data curation, Formal analysis, Visualization, Project administration, Writing -- original draft, and Writing -- review and editing.

\medskip

\noindent\textbf{Lynn Huestegge:}
Conceptualization, Methodology, Supervision, Funding acquisition, and Writing -- review and editing.

\medskip

\noindent\textbf{Anne Böckler-Raettig:}
Conceptualization, Methodology, Supervision, Funding acquisition, and Writing -- review and editing.

\medskip

All authors reviewed and approved the final manuscript.

\section*{Funding}

This work was funded by the Deutsche Forschungsgemeinschaft (DFG, German Research Foundation) under project number 562993814.

\section*{Ethics statement}

This methodological study used bibliographic records and screening outputs produced during an evidence-synthesis project. No patient data or directly identifiable personal data were collected or analysed. Formal ethics approval was not required for this study.

\section*{Data availability}

The preregistration for this study is available through the Open Science Framework at \url{https://osf.io/cnsr2/overview?view_only=697b0ae9f0234f069bc18173626ba954}.

The screening manual, prompts, input data, screening outputs, detailed results, and analysis code supporting the findings are available through the Open Science Framework at
\url{https://osf.io/v2q4r}.

\subsection*{Generative AI use}

Generative AI tools were used both as part of the research methodology and during manuscript preparation. As described in the Methods, OpenAI ChatGPT 5.4 Thinking, Google Gemini 3 Thinking, and Google Gemini 3.1 Pro were used to perform the LLM-based screening workflows evaluated in this study. Full records of the prompts and methodological procedures used for these research workflows were retained and are provided in the associated supplementary materials and research repository.

During manuscript preparation, OpenAI ChatGPT 5.4 and ChatGPT 5.6 were used for language editing and rephrasing, restructuring and shortening sections, assistance with LaTeX, checking analyses, and assistance with the development of R code. ChatGPT 5.4 also assisted with development of the R code used to generate Figure~1; the figure itself was generated from the study data using R and was not created using an AI image-generation model.

The authors reviewed and verified all AI-assisted text, code, analyses, references, and other outputs for accuracy and originality. The terms of use of the AI tools were reviewed and considered suitable for the uses described above and for publication. The authors take full responsibility for the integrity and accuracy of the entire manuscript, including its references.

\section*{Related work}

This manuscript contains original unpublished work and is not under consideration for publication elsewhere. The methodological study was embedded in the scoping review reported separately by Figalová et al. The companion report presents the substantive findings of the scoping review, whereas the present manuscript evaluates the human and LLM-based screening workflows. The relationship between the two reports and any overlapping methods or data is disclosed in the manuscript.


\bibliographystyle{apalike}
\bibliography{sample-base}

@article{tricco2018prisma,
  author  = {Tricco, Andrea C. and Lillie, Erin and Zarin, Wasifa and O'Brien, Kelly K. and Colquhoun, Heather and Levac, Danielle and Moher, David and Peters, Micah D. J. and Horsley, Tanya and Weeks, Laura and Hempel, Susanne and Akl, Elie A. and Chang, Christine and McGowan, Jessie and Stewart, Lesley and Hartling, Lisa and Aldcroft, Adrian and Wilson, Michael G. and Garritty, Chantelle and Lewin, Simon and Godfrey, Christina M. and Macdonald, Marilyn T. and Langlois, Etienne V. and Soares-Weiser, Karla and Moriarty, Jo and Clifford, Tammy and Tun{\c{c}}alp, {\"O}zge and Straus, Sharon E.},
  title   = {PRISMA Extension for Scoping Reviews (PRISMA-ScR): Checklist and Explanation},
  journal = {Annals of Internal Medicine},
  year    = {2018},
  volume  = {169},
  number  = {7},
  pages   = {467--473},
  doi     = {10.7326/M18-0850}
}

@article{vandeschoot2021asreview,
  author  = {van de Schoot, Rens and de Bruin, Jonathan and Schram, Raoul and Zahedi, Parisa and de Boer, Jan and Weijdema, Felix and Kramer, Bianca and Huijts, Martijn and Hoogerwerf, Maarten and Ferdinands, Gerbrich and Harkema, Albert and Willemsen, Joukje and Ma, Yongchao and Fang, Qixiang and Hindriks, Sybren and Tummers, Lars and Oberski, Daniel L.},
  title   = {An open source machine learning framework for efficient and transparent systematic reviews},
  journal = {Nature Machine Intelligence},
  year    = {2021},
  volume  = {3},
  number  = {2},
  pages   = {125--133},
  doi     = {10.1038/s42256-020-00287-7}
}

@article{homiar2025development,
  author  = {Homiar, Ava and Thomas, James and Ostinelli, Edoardo G. and Kennett, Jaycee and Friedrich, Claire and Cuijpers, Pim and Harrer, Mathias and Leucht, Stefan and Miguel, Clara and Rodolico, Alessandro and Kataoka, Yuki and Takayama, Tomohiro and Yoshimura, Keisuke and So, Ryuhei and Tsujimoto, Yasushi and Yamagishi, Yosuke and Takagi, Shiro and Sakata, Masatsugu and Ba{\v{s}}i{\'c}, {\DJ}or{\dj}e and others},
  title   = {Development and evaluation of prompts for a large language model to screen titles and abstracts in a living systematic review},
  journal = {BMJ Mental Health},
  year    = {2025},
  volume  = {28},
  number  = {1},
  pages   = {e301762},
  doi     = {10.1136/bmjment-2025-301762}
}

@article{landschaft2024implementation,
  author  = {Landschaft, Assaf and Antweiler, Dario and Mackay, Sina and Kugler, Sabine and R{\"u}ping, Stefan and Wrobel, Stefan and H{\"o}res, Timm and Allende-Cid, H{\'e}ctor},
  title   = {Implementation and evaluation of an additional GPT-4-based reviewer in PRISMA-based medical systematic literature reviews},
  journal = {International Journal of Medical Informatics},
  year    = {2024},
  volume  = {189},
  pages   = {105531},
  doi     = {10.1016/j.ijmedinf.2024.105531}
}

@misc{shailendra2025lprisma,
  author       = {Shailendra, Samar and Kadel, Rajan and Sharma, Aakanksha and Tahidul, Islam Mohammad and Saxena, Urvashi Rahul},
  title        = {L-PRISMA: An Extension of PRISMA in the Era of Generative Artificial Intelligence (GenAI)},
  year         = {2026},
  howpublished = {TechRxiv preprint},
  doi          = {10.36227/techrxiv.177040570.06710783/v1},
  note         = {Preprint; posted February 6, 2026}
}

@article{harasgama2026artificial,
  author  = {Harasgama, Sashika and Pearce, Helen and Appel, Cameron and Loftus, Liam and Painter, Helena and Kuhn, Isla and Karpusheff, Justine and Ceesay, Aji and Ford, John},
  title   = {Artificial Intelligence Tools for Automating Evidence Synthesis: Scoping Review},
  journal = {Journal of Medical Internet Research},
  year    = {2026},
  volume  = {28},
  pages   = {e81597},
  doi     = {10.2196/81597}
}

@article{lieberum2025large,
  author  = {Lieberum, Judith-Lisa and T{\"o}ws, Markus and Metzendorf, Maria-Inti and Heilmeyer, Felix and Siemens, Waldemar and Haverkamp, Christian and B{\"o}hringer, Daniel and Meerpohl, Joerg J. and Eisele-Metzger, Angelika},
  title   = {Large language models for conducting systematic reviews: On the rise, but not yet ready for use---a scoping review},
  journal = {Journal of Clinical Epidemiology},
  year    = {2025},
  volume  = {181},
  pages   = {111746},
  doi     = {10.1016/j.jclinepi.2025.111746}
}

@article{njei2026artificial,
  author  = {Njei, Basile and Al-Ajlouni, Yazan A. and Sidney Kanmounye, Ulrick and Boateng, Samuel and Loic Nguefang, Guy and Njei, Ngwa and others},
  title   = {Artificial intelligence agents in healthcare research: A scoping review},
  journal = {PLOS ONE},
  year    = {2026},
  volume  = {21},
  number  = {2},
  pages   = {e0342182},
  doi     = {10.1371/journal.pone.0342182}
}

@misc{cao2026automation,
  author       = {Cao, Christian and Arora, Rohit and Cento, Paul and Budak, Adil and Manta, Katherine and Farahani, Elina and Cecere, Matthew and Selemon, Anabel and Sang, Jason and Gong, Ling Xi and Kloosterman, Robert and Jiang, Scott and Saleh, Richard and Margalik, Denis and Lin, James and Jomy, Jane and Xie, Jerry and Chen, David and Gorla, Jaswanth and Lee, Sylvia and Zhang, Kelvin and Kuang, Jennifer and Ware, Harriet and Whelan, Mairead and Teja, Bijan and Leung, Alexander A. and Arora, Rahul K. and Pillay, Jennifer and Hartling, Lisa and Noetel, Michael and Emerson, David B. and Detsky, Allan S. and Tricco, Andrea C. and Church, George M. and Moher, David and Bobrovitz, Niklas},
  title        = {Automation of Systematic Reviews with Large Language Models},
  year         = {2026},
  howpublished = {medRxiv preprint},
  doi          = {10.1101/2025.06.13.25329541},
  note         = {Preprint; latest version checked: version 4, May 4, 2026}
}

@article{galli2025large,
  author  = {Galli, Carlo and Gavrilova, Anna V. and Calciolari, Elena},
  title   = {Large Language Models in Systematic Review Screening: Opportunities, Challenges, and Methodological Considerations},
  journal = {Information},
  year    = {2025},
  volume  = {16},
  number  = {5},
  pages   = {378},
  doi     = {10.3390/info16050378}
}

@misc{wilkins2023automated,
  author        = {Wilkins, David},
  title         = {Automated title and abstract screening for scoping reviews using the GPT-4 Large Language Model},
  year          = {2023},
  howpublished  = {arXiv preprint},
  eprint        = {2311.07918},
  archivePrefix = {arXiv},
  doi           = {10.48550/arXiv.2311.07918},
  note          = {Preprint}
}

@article{li2024evaluating,
  author  = {Li, Michael and Sun, Jianping and Tan, Xianming},
  title   = {Evaluating the effectiveness of large language models in abstract screening: A comparative analysis},
  journal = {Systematic Reviews},
  year    = {2024},
  volume  = {13},
  number  = {1},
  pages   = {219},
  doi     = {10.1186/s13643-024-02609-x}
}

@article{syriani2024screening,
  author  = {Syriani, Eugene and David, Istvan and Kumar, Gauransh},
  title   = {Screening articles for systematic reviews with ChatGPT},
  journal = {Journal of Computer Languages},
  year    = {2024},
  volume  = {80},
  pages   = {101287},
  doi     = {10.1016/j.cola.2024.101287}
}

@inproceedings{huotala2024promise,
  author    = {Huotala, Aleksi and Kuutila, Miikka and Ralph, Paul and M{\"a}ntyl{\"a}, Mika},
  title     = {The Promise and Challenges of Using LLMs to Accelerate the Screening Process of Systematic Reviews},
  booktitle = {Proceedings of the 28th International Conference on Evaluation and Assessment in Software Engineering},
  series    = {EASE 2024},
  year      = {2024},
  pages     = {273--283},
  publisher = {Association for Computing Machinery},
  address   = {New York, NY, USA},
  doi       = {10.1145/3661167.3661172}
}

@misc{krag2024large,
  author       = {Krag, Christian H. and Balschmidt, Signe and others},
  title        = {Large language models for abstract screening in systematic- and scoping reviews: A diagnostic test accuracy study},
  year         = {2024},
  howpublished = {medRxiv preprint},
  doi          = {10.1101/2024.10.01.24314702},
  note         = {Preprint}
}

@article{sciurti2026compact,
  author  = {Sciurti, Antonio and Migliara, Giuseppe and Siena, Leonardo Maria and Isonne, Claudia and De Blasiis, Maria Roberta and Sinopoli, Alessandra and Iera, Jessica and Marzuillo, Carolina and De Vito, Corrado and Villari, Paolo and Baccolini, Valentina},
  title   = {Compact large language models for title and abstract screening in systematic reviews: An assessment of feasibility, accuracy, and workload reduction},
  journal = {Research Synthesis Methods},
  year    = {2026},
  volume  = {17},
  number  = {2},
  pages   = {332--347},
  doi     = {10.1017/rsm.2025.10044},
  note    = {Published online 13 November 2025}
}

@article{akinseloyin2026multiagent,
  author  = {Akinseloyin, Opeoluwa and Jiang, Xiaorui and Palade, Vasile},
  title   = {Large language model-based multiagent collaboration for abstract screening toward automated systematic reviews},
  journal = {Biology Methods and Protocols},
  year    = {2026},
  volume  = {11},
  number  = {1},
  pages   = {bpag006},
  doi     = {10.1093/biomethods/bpag006}
}

@misc{wu2026prism,
  author        = {Wu, Yuhe and Wang, Guangyu and Chen, Yuran and Zhang, Jiatong and Zhang, Yutong and Chen, Yujie and Shang, Jiaming and Zhang, Guang and Liu, Zhuang},
  title         = {PRISM: Probing Reasoning, Instruction, and Source Memory in LLM Hallucinations},
  year          = {2026},
  howpublished  = {arXiv preprint},
  eprint        = {2604.16909},
  archivePrefix = {arXiv},
  doi           = {10.48550/arXiv.2604.16909},
  note          = {Preprint}
}

@misc{madeyski2025llm4screenlit,
  author        = {Madeyski, Lech and Kitchenham, Barbara and Shepperd, Martin},
  title         = {LLM4SCREENLIT: Recommendations on Assessing the Performance of Large Language Models for Screening Literature in Systematic Reviews},
  year          = {2025},
  howpublished  = {arXiv preprint},
  eprint        = {2511.12635},
  archivePrefix = {arXiv},
  doi           = {10.48550/arXiv.2511.12635},
  note          = {Preprint}
}

@misc{susnjak2023prisma,
  author        = {Susnjak, Teo},
  title         = {PRISMA-DFLLM: An Extension of PRISMA for Systematic Literature Reviews using Domain-specific Finetuned Large Language Models},
  year          = {2023},
  howpublished  = {arXiv preprint},
  eprint        = {2306.14905},
  archivePrefix = {arXiv},
  doi           = {10.48550/arXiv.2306.14905},
  note          = {Preprint}
}

@article{holst2025prismatraice,
  author  = {Holst, Dirk and Moenck, Keno and Koch, Julian and Schmedemann, Ole and Sch{\"u}ppstuhl, Thorsten},
  title   = {Transparent Reporting of AI in Systematic Literature Reviews: Development of the PRISMA-trAIce Checklist},
  journal = {JMIR AI},
  year    = {2025},
  volume  = {4},
  pages   = {e80247},
  doi     = {10.2196/80247}
}

@article{gallifant2025tripodllm,
  author  = {Gallifant, Jack and Afshar, Majid and Ameen, Saleem and Aphinyanaphongs, Yindalon and Chen, Shan and Cacciamani, Giovanni and Demner-Fushman, Dina and Dligach, Dmitriy and Daneshjou, Roxana and Fernandes, Corey and Hansen, Line H. and Landman, Adam and Lehmann, Lillian and McCoy, Liam G. and Miller, Tim and Moreno, Andres and Munch, Niels and Restrepo, Daniel and Savova, Guergana and Umeton, Renato and Gichoya, Judy Wawira and Collins, Gary S. and Moons, Karel G. M. and Celi, Leo Anthony and Bitterman, Danielle S.},
  title   = {The TRIPOD-LLM reporting guideline for studies using large language models},
  journal = {Nature Medicine},
  year    = {2025},
  volume  = {31},
  number  = {1},
  pages   = {60--69},
  doi     = {10.1038/s41591-024-03425-5}
}

@article{luo2025gamer,
  author  = {Luo, Xufei and Tham, Yih Chung and Giuffr{\`e}, Mauro and Ranisch, Robert and Daher, Mohammad and Lam, Kyle and Eriksen, Alexander Viktor and Hsu, Che-Wei and Ozaki, Akihiko and de Moraes, Fabio Ynoe and Khanna, Sahil and Su, Kuan-Pin and Begagi{\'c}, Emir and Bian, Zhaoxiang and Chen, Yiqun and Estill, Jan and GAMER Working Group},
  title   = {Reporting guideline for the use of Generative Artificial intelligence tools in MEdical Research: The GAMER Statement},
  journal = {BMJ Evidence-Based Medicine},
  year    = {2025},
  volume  = {30},
  number  = {6},
  pages   = {390--400},
  doi     = {10.1136/bmjebm-2025-113825}
}

@misc{figalova_bockler_huestegge_2026,
  title     = {Gaze semantics: A scoping review of how gaze conveys and constrains meaning in social contexts},
  author    = {Figalov{\'a}, Nikol and B{\"o}ckler, Anne and Huestegge, Lynn},
  year      = {2026},
  month     = jul,
  publisher = {PsyArXiv},
  doi       = {10.31234/osf.io/58hdw_v2},
  url       = {https://doi.org/10.31234/osf.io/58hdw_v2},
  note      = {Preprint}
}

@article{hanfstingl2024detecting,
  title={Detecting jingle and jangle fallacies by identifying consistencies and variabilities in study specifications--a call for research},
  author={Hanfstingl, Barbara and Oberleiter, Sandra and Pietschnig, Jakob and Tran, Ulrich S and Voracek, Martin},
  journal={Frontiers in Psychology},
  volume={15},
  pages={1404060},
  year={2024},
  publisher={Frontiers Media SA}
}

@article{sokolova2009systematic,
  title   = {A systematic analysis of performance measures for classification tasks},
  author  = {Sokolova, Marina and Lapalme, Guy},
  journal = {Information Processing \& Management},
  volume  = {45},
  number  = {4},
  pages   = {427--437},
  year    = {2009},
  doi     = {10.1016/j.ipm.2009.03.002}
}

@article{whiting2011quadas,
  title   = {{QUADAS-2}: A revised tool for the quality assessment of diagnostic accuracy studies},
  author  = {Whiting, Penny F. and Rutjes, Anne W. S. and Westwood, Marie E. and Mallett, Susan and Deeks, Jonathan J. and Reitsma, Johannes B. and Leeflang, Mariska M. G. and Sterne, Jonathan A. C. and Bossuyt, Patrick M. M.},
  journal = {Annals of Internal Medicine},
  volume  = {155},
  number  = {8},
  pages   = {529--536},
  year    = {2011},
  doi     = {10.7326/0003-4819-155-8-201110180-00009}
}

@article{belur2021interrater,
  title   = {Interrater Reliability in Systematic Review Methodology: Exploring Variation in Coder Decision-Making},
  author  = {Belur, Jyoti and Tompson, Lisa and Thornton, Amy and Simon, Miranda},
  journal = {Sociological Methods \& Research},
  volume  = {50},
  number  = {2},
  pages   = {837--865},
  year    = {2021},
  doi     = {10.1177/0049124118799372}
}

@article{wang2020error,
  title   = {Error Rates of Human Reviewers during Abstract Screening in Systematic Reviews},
  author  = {Wang, Zhen and Nayfeh, Tarek and Tetzlaff, Jennifer and O'Blenis, Peter and Murad, Mohammad Hassan},
  journal = {PLOS ONE},
  volume  = {15},
  number  = {1},
  pages   = {e0227742},
  year    = {2020},
  doi     = {10.1371/journal.pone.0227742}
}

@article{tversky1974judgment,
  title   = {Judgment under Uncertainty: Heuristics and Biases},
  author  = {Tversky, Amos and Kahneman, Daniel},
  journal = {Science},
  volume  = {185},
  number  = {4157},
  pages   = {1124--1131},
  year    = {1974},
  doi     = {10.1126/science.185.4157.1124}
}

@article{byrt1993bias,
  author  = {Byrt, Ted and Bishop, Janet and Carlin, John B.},
  title   = {Bias, Prevalence and Kappa},
  journal = {Journal of Clinical Epidemiology},
  year    = {1993},
  volume  = {46},
  number  = {5},
  pages   = {423--429},
  doi     = {10.1016/0895-4356(93)90018-V}
}

@article{gwet2008computing,
  author  = {Gwet, Kilem Li},
  title   = {Computing Inter-Rater Reliability and Its Variance in the
             Presence of High Agreement},
  journal = {British Journal of Mathematical and Statistical Psychology},
  year    = {2008},
  volume  = {61},
  number  = {1},
  pages   = {29--48},
  doi     = {10.1348/000711006X126600}
}

@article{peters2020updated,
  author  = {Peters, Micah D. J. and Marnie, Casey and Tricco, Andrea C. and Pollock, Danielle and Munn, Zachary and Alexander, Lyndsay and McInerney, Patricia and Godfrey, Christina M. and Khalil, Hanan},
  title   = {Updated Methodological Guidance for the Conduct of Scoping Reviews},
  journal = {JBI Evidence Synthesis},
  year    = {2020},
  volume  = {18},
  number  = {10},
  pages   = {2119--2126},
  doi     = {10.11124/JBIES-20-00167}
}

@article{dennstadt2024title,
  author  = {Dennst{\"a}dt, Fabio and Zink, Johannes and Putora, Paul Martin and Hastings, Janna and Cihoric, Nikola},
  title   = {Title and Abstract Screening for Literature Reviews Using Large Language Models: An Exploratory Study in the Biomedical Domain},
  journal = {Systematic Reviews},
  year    = {2024},
  volume  = {13},
  pages   = {158},
  doi     = {10.1186/s13643-024-02575-4}
}

@article{waffenschmidt2019single,
  author  = {Waffenschmidt, Siw and Knelangen, Marco and Sieben, Wiebke and B{\"u}hn, Stefanie and Pieper, Dawid},
  title   = {Single Screening versus Conventional Double Screening for Study Selection in Systematic Reviews: A Methodological Systematic Review},
  journal = {BMC Medical Research Methodology},
  year    = {2019},
  volume  = {19},
  pages   = {132},
  doi     = {10.1186/s12874-019-0782-0}
}

@article{sanghera2025high,
  author  = {Sanghera, Rohan and Thirunavukarasu, Arun James and El Khoury, Marc and O'Logbon, Jessica and Chen, Yuqing and Watt, Archie and Mahmood, Mustafa and Butt, Hamid and Nishimura, George and Soltan, Andrew A. S.},
  title   = {High-Performance Automated Abstract Screening with Large Language Model Ensembles},
  journal = {Journal of the American Medical Informatics Association},
  year    = {2025},
  volume  = {32},
  number  = {5},
  pages   = {893--904},
  doi     = {10.1093/jamia/ocaf050}
}

@article{fagerberg2026batch,
  author  = {Fagerberg, Petter and Sallander, Oscar and Patil, Kim Vikhe and Berg, Anders and Nyman, Anastasia and Borg, Natalia and Lind{\'e}n, Thomas},
  title   = {Batch Size Effects on Mid-2025 State-of-the-Art Large Language Model Performance in Automated Title and Abstract Screening},
  journal = {Cochrane Evidence Synthesis and Methods},
  year    = {2026},
  volume  = {4},
  number  = {3},
  pages   = {e70082},
  doi     = {10.1002/cesm.70082}
}

@article{hilkenmeier2026full,
  author  = {Hilkenmeier, Frederic and Stoltenberg, Merle and Stierle, Christian},
  title   = {Using Full Agreement across Multiple Large Language Models for Title-and-Abstract Screening in Systematic Reviews: A Proof-of-Concept},
  journal = {Systematic Reviews},
  year    = {2026},
  volume  = {15},
  pages   = {191},
  doi     = {10.1186/s13643-026-03228-4}
}

@article{laignelot2026large,
  author  = {Laignelot, Florian and Martin, Guillaume L. and Ossman, Mohamad and Pingeon, Oph{\'e}lie and Boubaker, Amine and Picovschi, Emma and Kim, Jia and Tannier, Xavier and Cohen, J{\'e}r{\'e}mie F. and Dechartres, Agn{\`e}s},
  title   = {Large Language Models Show Promising Performance for Some Systematic Review Tasks but Call for Cautious Implementation: A Systematic Review},
  journal = {Journal of Clinical Epidemiology},
  year    = {2026},
  volume  = {194},
  pages   = {112221},
  doi     = {10.1016/j.jclinepi.2026.112221}
}



\begin{table}[htbp]
\centering
\caption{LLM-based title-and-abstract screening workflows evaluated in the study. Each run comprised one complete pass through all 1,131 benchmark records.}
\label{tab:llm_runs}
\footnotesize
\setlength{\tabcolsep}{4pt}
\renewcommand{\arraystretch}{1.08}

\begin{tabularx}{\linewidth}{
@{}
p{0.21\linewidth}
p{0.16\linewidth}
p{0.27\linewidth}
X
@{}
}
\toprule
\textbf{Model}
& \textbf{Date}
& \textbf{Processing configuration}
& \textbf{Role in study} \tabularnewline
\midrule

ChatGPT 5.4 Thinking
& March 17, 2026
& Files of approx. 100 records
& Advancement workflow \tabularnewline

ChatGPT 5.4 Thinking
& March 17, 2026
& All 1,131 records at once
& Comparative run \tabularnewline

ChatGPT 5.4 Thinking
& March 18, 2026
& Files of approx. 100 records
& Nominally identical repeat of the March 17 file-batch workflow \tabularnewline

ChatGPT 5.4 Thinking
& March 19--24, 2026
& Sequential groups of 10 records
& Exploratory comparative run \tabularnewline

Gemini 3 Thinking
& March 17, 2026
& Files of approx. 100 records
& Advancement workflow \tabularnewline

Gemini 3 Thinking
& March 17, 2026
& All 1,131 records at once
& Comparative run \tabularnewline

Gemini 3.1 Pro
& March 18, 2026
& Files of approx. 100 records
& Comparative run \tabularnewline

\bottomrule
\end{tabularx}

\vspace{3pt}
\begin{minipage}{\linewidth}
\footnotesize
\RaggedRight
\textit{Note.} Model names are reported exactly as displayed in the hosted interfaces at the time of data collection. The March 18 ChatGPT file-batch run repeated the March 17 procedure using the same displayed model, prompt, record order, file structure, and interaction procedure; undocumented changes to the hosted system cannot be excluded. The sequential groups-of-10 configuration constituted one complete run conducted across several days.
\end{minipage}
\end{table}
\clearpage


\begin{table}[htbp]
\centering
\caption{Evaluation dimensions, outcome measures, and analytical sets used to address the research questions.}
\label{tab:outcome_measures_overview}

\footnotesize
\setlength{\tabcolsep}{5pt}
\renewcommand{\arraystretch}{1.12}

\begin{tabularx}{\linewidth}{
@{}
L{0.07\linewidth}
L{0.20\linewidth}
Y
L{0.19\linewidth}
@{}
}
\toprule
\textbf{RQ}
& \textbf{Evaluation dimension}
& \textbf{Outcome measures}
& \textbf{Analytical set} \tabularnewline
\midrule

\textbf{RQ1}
& Recovery and retained workload
& Operational recall and missed eligible records quantified recovery of verified eligible records. Retained workload quantified the number of benchmark records retained for possible full-text assessment.
& 316 verified eligible records; all 1,131 benchmark records
\tabularnewline
\addlinespace[3pt]

\textbf{RQ2}
& Agreement between screening outputs
& Overall agreement, Cohen's $\kappa$, Gwet's AC1, and positive and negative agreement compared binary retained/not-retained decisions.
& All 1,131 benchmark records
\tabularnewline
\addlinespace[3pt]

\textbf{RQ3}
& Conditional classification performance
& Precision, specificity, and $F_1$ described classification performance among records with verified full-text outcomes.
& 859 full-text-assessed records
\tabularnewline
\addlinespace[3pt]

\textbf{RQ4}
& Run-to-run consistency
& The two nominally identical GPT-5.4 file-batch runs were compared using record-level agreement, retained workload, operational recall, missed eligible records, and discordant decisions.
& All 1,131 benchmark records, including 316 verified eligible records
\tabularnewline
\addlinespace[3pt]

\textbf{RQ5}
& Missed-record patterns and retrospective checks
& Record-level analyses examined overlap in missed eligible records, unique recovery, and records that alternative procedures would have advanced. Thirteen non-advanced records retained by the later sequential run were reassessed post hoc.
& All 1,131 benchmark records, including 316 verified eligible records; 13 reassessed records
\tabularnewline
\addlinespace[3pt]

\textbf{Explor.}
& Pairwise workflow complementarity
& A liberal union rule quantified recovery and retained workload when a record was retained if either member of a workflow pair retained it.
& All 1,131 benchmark records, including 316 verified eligible records
\tabularnewline
\addlinespace[3pt]

\textbf{Suppl.}
& Procedural burden
& Human screening time and throughput and available LLM timing, interaction, and conversation-level measures were summarised descriptively.
& Available process logs
\tabularnewline

\bottomrule
\end{tabularx}

\vspace{3pt}
\begin{minipage}{\linewidth}
\footnotesize
\RaggedRight
\textit{Note.} RQ = research question. Operational recall was calculated against the 316 verified eligible records and retained workload against the complete 1,131-record benchmark. Precision, specificity, and $F_1$ were conditional on the 859 records assessed at full text and should not be interpreted as complete-benchmark classification estimates.
\end{minipage}
\end{table}
\clearpage


\begin{table}[htbp]
\centering
\caption{Retained workload and recovery of verified eligible records for each screening workflow.}
\label{tab:workload_recall}
\footnotesize
\setlength{\tabcolsep}{3pt}
\renewcommand{\arraystretch}{0.98}

\begin{tabularx}{\linewidth}{@{}
>{\raggedright\arraybackslash}p{0.24\linewidth}
>{\raggedleft\arraybackslash}p{0.16\linewidth}
>{\raggedleft\arraybackslash}p{0.16\linewidth}
>{\raggedleft\arraybackslash}p{0.15\linewidth}
>{\raggedleft\arraybackslash}p{0.23\linewidth}
@{}}
\toprule
\textbf{Workflow}
& \shortstack{\textbf{Retained}\\\textbf{workload,}\\\textbf{$n$ (\%)}}
& \shortstack{\textbf{Eligible}\\\textbf{retained,}\\\textbf{$n$ (\%)}}
& \shortstack{\textbf{Eligible}\\\textbf{missed,}\\\textbf{$n$ (\%)}}
& \shortstack{\textbf{Operational recall}\\\textbf{(95\% CI)}} \tabularnewline
\midrule

Single reviewer
& 477 (42.2\%)
& 262 (82.9\%)
& 54 (17.1\%)
& 0.829 (0.785--0.870) \tabularnewline

Distributed team
& 509 (45.0\%)
& 260 (82.3\%)
& 56 (17.7\%)
& 0.823 (0.778--0.864) \tabularnewline

Gemini 3: file batches
& 714 (63.1\%)
& 244 (77.2\%)
& 72 (22.8\%)
& 0.772 (0.725--0.816) \tabularnewline

Gemini 3: all at once
& 667 (59.0\%)
& 209 (66.1\%)
& 107 (33.9\%)
& 0.661 (0.608--0.715) \tabularnewline

GPT-5.4: file batches
& 493 (43.6\%)
& 261 (82.6\%)
& 55 (17.4\%)
& 0.826 (0.785--0.867) \tabularnewline

GPT-5.4: all at once
& 391 (34.6\%)
& 211 (66.8\%)
& 105 (33.2\%)
& 0.668 (0.614--0.718) \tabularnewline

Gemini 3.1: file batches
& 641 (56.7\%)
& 265 (83.9\%)
& 51 (16.1\%)
& 0.839 (0.797--0.877) \tabularnewline

GPT-5.4: file-batch rerun
& 509 (45.0\%)
& 262 (82.9\%)
& 54 (17.1\%)
& 0.829 (0.788--0.870) \tabularnewline

GPT-5.4: batches of 10
& 503 (44.5\%)
& 252 (79.7\%)
& 64 (20.3\%)
& 0.797 (0.753--0.842) \tabularnewline

\bottomrule
\end{tabularx}

\vspace{3pt}
\begin{minipage}{\linewidth}
\footnotesize
\RaggedRight
\textit{Note.} Retained workload was calculated across all 1,131 benchmark records. Eligible retained, eligible missed, and operational recall were calculated against the 316 records judged eligible after full-text assessment. Operational recall is therefore an estimate against the verified eligible set rather than complete-benchmark recall. Confidence intervals are 95\% percentile intervals from 10,000 stratified bootstrap resamples.
\end{minipage}
\end{table}
\clearpage


\begin{table}[htbp]
\centering
\caption{Selected pairwise agreement comparisons for binary retained/not-retained decisions across the 1,131-record benchmark.}
\label{tab:main_agreement_summary}
\scriptsize
\setlength{\tabcolsep}{2.5pt}
\renewcommand{\arraystretch}{1.08}
\providecommand{\pairindent}{\hspace*{1.2em}}

\begin{tabularx}{\linewidth}{@{}
>{\raggedright\arraybackslash}p{0.27\linewidth}
>{\centering\arraybackslash}p{0.13\linewidth}
>{\centering\arraybackslash}p{0.14\linewidth}
>{\centering\arraybackslash}p{0.14\linewidth}
>{\centering\arraybackslash}p{0.13\linewidth}
>{\centering\arraybackslash}p{0.13\linewidth}
@{}}
\toprule
\textbf{Workflow pair}
& \shortstack{\textbf{Agreement, \%}\\\textbf{(95\% CI)}}
& \shortstack{\textbf{Positive}\\\textbf{agreement}\\\textbf{(95\% CI)}}
& \shortstack{\textbf{Negative}\\\textbf{agreement}\\\textbf{(95\% CI)}}
& \shortstack{\textbf{Cohen's}\\\textbf{$\kappa$}\\\textbf{(95\% CI)}}
& \shortstack{\textbf{Gwet's}\\\textbf{AC1}\\\textbf{(95\% CI)}} \tabularnewline
\midrule

\multicolumn{6}{@{}l}{\textit{Human--human}} \tabularnewline

\pairindent Single reviewer vs distributed team
& \shortstack{72.8\\(70.2--75.3)}
& \shortstack{0.688\\(0.653--0.720)}
& \shortstack{0.759\\(0.732--0.784)}
& \shortstack{0.447\\(0.393--0.499)}
& \shortstack{0.464\\(0.413--0.516)} \tabularnewline
\addlinespace[2pt]

\multicolumn{6}{@{}l}{\textit{Single reviewer vs LLM}} \tabularnewline

\shortstack[l]{\pairindent Single reviewer vs Gemini 3 file\\\pairindent batches}
& \shortstack{55.0\\(52.1--57.8)}
& \shortstack{0.573\\(0.539--0.605)}
& \shortstack{0.525\\(0.488--0.560)}
& \shortstack{0.135\\(0.083--0.187)}
& \shortstack{0.102\\(0.045--0.160)} \tabularnewline

\shortstack[l]{\pairindent Single reviewer vs GPT-5.4 file\\\pairindent batches}
& \shortstack{71.2\\(68.4--73.8)}
& \shortstack{0.664\\(0.629--0.698)}
& \shortstack{0.748\\(0.720--0.773)}
& \shortstack{0.412\\(0.356--0.465)}
& \shortstack{0.435\\(0.380--0.488)} \tabularnewline
\addlinespace[2pt]

\multicolumn{6}{@{}l}{\textit{Distributed team vs LLM}} \tabularnewline

\shortstack[l]{\pairindent Distributed team vs Gemini 3 file\\\pairindent batches}
& \shortstack{55.7\\(52.7--58.6)}
& \shortstack{0.590\\(0.556--0.623)}
& \shortstack{0.518\\(0.479--0.554)}
& \shortstack{0.137\\(0.081--0.190)}
& \shortstack{0.120\\(0.060--0.180)} \tabularnewline

\shortstack[l]{\pairindent Distributed team vs GPT-5.4 file\\\pairindent batches}
& \shortstack{74.0\\(71.4--76.5)}
& \shortstack{0.707\\(0.673--0.737)}
& \shortstack{0.767\\(0.740--0.791)}
& \shortstack{0.473\\(0.420--0.523)}
& \shortstack{0.487\\(0.434--0.537)} \tabularnewline
\addlinespace[2pt]

\multicolumn{6}{@{}l}{\textit{Within-model processing-configuration comparisons}} \tabularnewline

\shortstack[l]{\pairindent Gemini 3 file batches vs Gemini 3\\\pairindent all at once}
& \shortstack{62.4\\(59.5--65.3)}
& \shortstack{0.692\\(0.663--0.720)}
& \shortstack{0.518\\(0.476--0.557)}
& \shortstack{0.211\\(0.153--0.268)}
& \shortstack{0.283\\(0.224--0.342)} \tabularnewline

\shortstack[l]{\pairindent GPT-5.4 file batches vs GPT-5.4\\\pairindent all at once}
& \shortstack{82.5\\(80.3--84.7)}
& \shortstack{0.776\\(0.745--0.805)}
& \shortstack{0.856\\(0.836--0.876)}
& \shortstack{0.635\\(0.590--0.680)}
& \shortstack{0.666\\(0.622--0.709)} \tabularnewline
\addlinespace[2pt]

\multicolumn{6}{@{}l}{\textit{Run-to-run consistency}} \tabularnewline

\shortstack[l]{\pairindent GPT-5.4 file batches vs GPT-5.4\\\pairindent file-batch rerun}
& \shortstack{91.7\\(90.0--93.2)}
& \shortstack{0.906\\(0.886--0.924)}
& \shortstack{0.925\\(0.909--0.940)}
& \shortstack{0.832\\(0.797--0.863)}
& \shortstack{0.836\\(0.802--0.867)} \tabularnewline

\bottomrule
\end{tabularx}

\vspace{3pt}
\begin{minipage}{\linewidth}
\footnotesize
\RaggedRight
\textit{Note.} Agreement was calculated for the binary retained/not-retained classification across all 1,131 benchmark records. Positive agreement quantifies concordance in retaining records; negative agreement quantifies concordance in not retaining records. Confidence intervals are 95\% percentile intervals from 10,000 multinomial resamples of the observed $2 \times 2$ agreement table for each workflow pair. Complete results for all 36 workflow pairs are provided in the Supplementary Materials.
\end{minipage}
\end{table}
\clearpage


\begin{table}[htbp]
\centering
\caption{Conditional classification performance among the 859 records with verified full-text outcomes.}
\label{tab:conditional_classification_results}
\footnotesize
\setlength{\tabcolsep}{2.5pt}
\renewcommand{\arraystretch}{1.08}

\begin{tabularx}{\linewidth}{@{}
>{\raggedright\arraybackslash}p{0.27\linewidth}
>{\centering\arraybackslash}p{0.06\linewidth}
>{\centering\arraybackslash}p{0.06\linewidth}
>{\centering\arraybackslash}p{0.18\linewidth}
>{\centering\arraybackslash}p{0.18\linewidth}
>{\centering\arraybackslash}p{0.17\linewidth}
@{}}
\toprule
\textbf{Workflow}
& \textbf{FP}
& \textbf{TN}
& \shortstack{\textbf{Precision}\\\textbf{(95\% CI)}}
& \shortstack{\textbf{Specificity}\\\textbf{(95\% CI)}}
& \shortstack{\textbf{$F_1$}\\\textbf{(95\% CI)}} \tabularnewline
\midrule

Single reviewer
& 178
& 365
& \shortstack{0.595\\(0.565--0.627)}
& \shortstack{0.672\\(0.634--0.711)}
& \shortstack{0.693\\(0.662--0.723)} \tabularnewline
\addlinespace[3pt]

Distributed team
& 208
& 335
& \shortstack{0.556\\(0.527--0.585)}
& \shortstack{0.617\\(0.575--0.657)}
& \shortstack{0.663\\(0.633--0.692)} \tabularnewline
\addlinespace[3pt]

Gemini 3: file batches
& 427
& 116
& \shortstack{0.364\\(0.346--0.381)}
& \shortstack{0.214\\(0.179--0.250)}
& \shortstack{0.494\\(0.470--0.518)} \tabularnewline
\addlinespace[3pt]

Gemini 3: all at once
& 335
& 208
& \shortstack{0.384\\(0.359--0.409)}
& \shortstack{0.383\\(0.343--0.424)}
& \shortstack{0.486\\(0.453--0.518)} \tabularnewline
\addlinespace[3pt]

GPT-5.4: file batches
& 202
& 341
& \shortstack{0.564\\(0.535--0.594)}
& \shortstack{0.628\\(0.587--0.669)}
& \shortstack{0.670\\(0.640--0.699)} \tabularnewline
\addlinespace[3pt]

GPT-5.4: all at once
& 153
& 390
& \shortstack{0.580\\(0.543--0.617)}
& \shortstack{0.718\\(0.681--0.755)}
& \shortstack{0.621\\(0.583--0.658)} \tabularnewline
\addlinespace[3pt]

Gemini 3.1: file batches
& 317
& 226
& \shortstack{0.455\\(0.434--0.477)}
& \shortstack{0.416\\(0.376--0.459)}
& \shortstack{0.590\\(0.565--0.615)} \tabularnewline
\addlinespace[3pt]

GPT-5.4: file-batch rerun
& 208
& 335
& \shortstack{0.557\\(0.529--0.587)}
& \shortstack{0.617\\(0.576--0.657)}
& \shortstack{0.667\\(0.638--0.696)} \tabularnewline
\addlinespace[3pt]

GPT-5.4: batches of 10
& 204
& 339
& \shortstack{0.553\\(0.523--0.584)}
& \shortstack{0.624\\(0.582--0.665)}
& \shortstack{0.653\\(0.622--0.683)} \tabularnewline

\bottomrule
\end{tabularx}

\vspace{3pt}
\begin{minipage}{\linewidth}
\footnotesize
\RaggedRight
\textit{Note.} Classification measures were calculated only among the 859 records assessed at full text, comprising 316 verified eligible and 543 ineligible records. FP = false positives, defined here as full-text-ineligible records retained during title-and-abstract screening; TN = true negatives, defined as full-text-ineligible records not retained during title-and-abstract screening. Recall is not repeated because it is numerically identical to the operational recall reported in Table~\ref{tab:workload_recall}. These estimates are conditional on full-text verification and are not complete-benchmark classification measures. Confidence intervals are 95\% percentile intervals from 10,000 stratified bootstrap resamples.
\end{minipage}
\end{table}
\clearpage


\begin{table}[htbp]
\centering
\caption{Logged human screening time and throughput during formal title-and-abstract screening.}
\label{tab:human_screening_time}
\footnotesize
\setlength{\tabcolsep}{6pt}
\renewcommand{\arraystretch}{1.08}

\begin{tabular}{lrrrr}
\toprule
\textbf{Workflow or reviewer}
& \textbf{Records screened}
& \textbf{Time, min}
& \textbf{Min per record}
& \textbf{Records per hour} \tabularnewline
\midrule

Single reviewer
& 1,131
& 1,176
& 1.04
& 57.7 \tabularnewline

Distributed team, total
& 1,131
& 1,628
& 1.44
& 41.7 \tabularnewline

Assistant 1
& 300
& 418
& 1.39
& 43.1 \tabularnewline

Assistant 2
& 331
& 505
& 1.53
& 39.3 \tabularnewline

Assistant 3
& 200
& 300
& 1.50
& 40.0 \tabularnewline

Assistant 4
& 300
& 405
& 1.35
& 44.4 \tabularnewline

\bottomrule
\end{tabular}

\vspace{3pt}
\begin{minipage}{\linewidth}
\footnotesize
\RaggedRight
\textit{Note.} Times refer to formal screening only. Calibration, training, breaks, and recovery time were excluded. Screening times for the single reviewer and assistants were self-reported in the screening sheets. Distributed-team time represents summed person-time across assistants rather than elapsed calendar time and is therefore not directly comparable with the elapsed processing times reported descriptively for the LLM workflows.
\end{minipage}
\end{table}
\clearpage

\end{document}